\pdfoutput=1

\documentclass[11pt]{article}

\usepackage[]{acl}

\usepackage{times}
\usepackage{adjustbox}
\usepackage{multirow}
\usepackage{latexsym}
\usepackage{amsmath}
\usepackage{amssymb}
\usepackage{makecell}

\usepackage[T1]{fontenc}

\usepackage[utf8]{inputenc}

\usepackage{microtype}

\title{Naive Bayes-based Context Extension for Large Language Models}

     \author{Jianlin Su\thanks{\hspace{1ex} These authors contributed equally to this work.} \quad 
	     	Murtadha Ahmed\footnotemark[1] \quad 
	     	Wenbo  \quad 
	     	Luo Ao\quad 
	     	Mingren Zhu \quad 
	     	Yunfeng Liu\\ 
	     	Zhuiyi Technology Co. Ltd., Shenzhen, Guangdong, China \\
	     	{\tt \{bojonesu,a.murtadha,brucewen,luoao,mingren,glenliu\}@wezhuiyi.com}
	     }

\begin{document}
\maketitle
\begin{abstract}
Large Language Models (LLMs) have shown promising in-context learning abilities. However, conventional  In-Context Learning (ICL) approaches are often impeded by length limitations of transformer architecture,
which pose challenges when attempting to effectively integrate supervision from a substantial number of demonstration examples. In this paper, we introduce a novel framework, called Naive Bayes-based Context Extension (NBCE), to enable existing LLMs to perform ICL with an increased number of demonstrations by significantly expanding their context size. Importantly, this expansion does not require fine-tuning or dependence on particular model architectures, all the while preserving linear efficiency.
NBCE initially splits the context into equal-sized windows fitting the target LLM's maximum length. Then, it introduces a voting mechanism to select the most relevant window, regarded as the posterior context. Finally, it employs Bayes' theorem to generate the test task. Our  experimental results demonstrate that NBCE substantially enhances performance, particularly as the number of demonstration examples increases, consistently outperforming alternative methods. The NBCE code will be made publicly accessible. The code NBCE is available at: \href{https://github.com/amurtadha/NBCE-master}{https://github.com/amurtadha/NBCE-master}
\end{abstract}

\section{Introduction} \label{sec:introduction}

Large Language Models (LLMs) have demonstrated remarkable capabilities in in-context learning (ICL), a paradigm that enables them to excel in various unseen tasks based on task examples or instructions within their context \cite{HanZDGLHQYZZHHJ21,Qiu-2003-08271}. Unlike traditional fine-tuning methods, ICL leverages  LLMs for downstream tasks solely through inference, eliminating the need for parameter updates and making it computationally efficient, bringing us closer to the goal of general AI. This approach has gained prominence as  LLMs continue to grow in scale \cite{BrownMRSKDNSSAA20,Zhang-2205-01068,Chowdhery-2204-02311}. 


The 2048-token context limit in popular  LLMs like GPT-3 poses challenges for scaling up ICL with more demonstration examples in ICL, due to architectural constraints and computational complexity.
Recent studies \cite{Garg0001TLV22,MinLZH22,ChenZZK022} improve ICL through meta-learning and fine-tuning on downstream tasks, but the limited diversity of annotated tasks and biases hinder generalization. 
Another line of research has explored various approaches to retraining long-range language models with extrapolation, extending them to 128 times the limit of existing  LLMs \cite{Mukai-2302-04931,Gu0WH23}. However, these approaches require additional training over several steps, which can be time-consuming.


Recently, \citet{Yaru-2212-06713} introduced structured prompting, encoding demonstrations with specific position embeddings for collective attention via a scaled mechanism. Extending this, \citet{pcw} proposed parallel context windows, utilizing individual encoding of examples with designed position and attention mechanisms. Addressing this issue is crucial for leveraging ICL effectively, especially in scenarios with ample examples.

In this paper, we introduce a novel framework called Naive Bayes-based Context Extension (NBCE) for large language models to significantly expand the number of demonstrations by orders of magnitude while greatly enhancing stability. Instead of simply merging all demonstrations, we partition the vast number of demonstrations into multiple groups, each independently processed by the language model. This approach ensures that the encoding complexity scales linearly with the number of groups, avoiding the quadratic complexity associated with considering all examples simultaneously. Following \citet{pcw,Yaru-2212-06713}, we align the position embeddings of grouped prompts to the right, placing them next to the test input. Subsequently, we leverage the Naive Bayes to encode the input by conditioning it on these grouped prompts. We conducted experiments across various tasks, including text classification, multi-choice, and open-ended tasks. NBCE effectively scales up the number of demonstrations, outperforming conventional in-context learning across different model sizes and tasks, while also significantly enhancing stability.

In brief, the contributions can be summarized as follows:
%

\begin{enumerate}
	\item We introduce an innovative framework known as Naive Bayes-based Context Extension (NBCE), designed to substantially increase the volume of demonstrations for large language models, thus enhancing stability on a significant scale.	
	\item We provide detailed technical insights to enable context expending of in-context learning tasks. The idea is to encode the test sample by conditioning it on a vast array of demonstrations sourced from the training dataset.	
	\item We conducted extensive experiments on benchmark NLP datasets, and our findings clearly highlight NBCE's remarkable capability to efficiently scale up the number of demonstrations, while significantly enhancing overall stability.
\end{enumerate}

\section{Approach}\label{sec:approach}


An example of our proposed NBCE is depicted in \ref{fig:nbce}.
Assume that we have a sequence, denoted as $T$, which we intend to generate. Furthermore, we have multiple relatively independent context sets, denoted as $S_1, S_2, \ldots, S_n$ (e.g., $n$ different paragraphs), each of which is sufficiently long and does not split a sentence into fragments. Suppose that the total length of these context sets exceeds the training length, but when combined with an individual $S_k$ and $T$, they still fall within the training length. Our objective is to generate $T$ based on the information contained in $S_1, S_2, \ldots, S_n$. In essence, we seek to estimate the conditional probability of $T$ given $S_1, S_2, \ldots, S_n$, which can be represented as $p(T|S_1,S_2,\ldots,S_n)$.



In straightforward terms, Naive Bayes can be understood as a combination of two key elements: Bayes' formula and an independence assumption:
\begin{equation}
p(T|S_1,S_2,\ldots,S_n) \propto p(S_1,S_2,\ldots,S_n|T)p(T),
\end{equation}
where, the symbol $\propto$ denotes proportionality, signifying that we are focusing solely on the relevant factors in a proportion while disregarding constant factors unrelated to the token sequence $T$.  This approach aligns with the underlying assumption of conditional independence:
\begin{figure}
	\centering
	\includegraphics[width=0.45\textwidth]{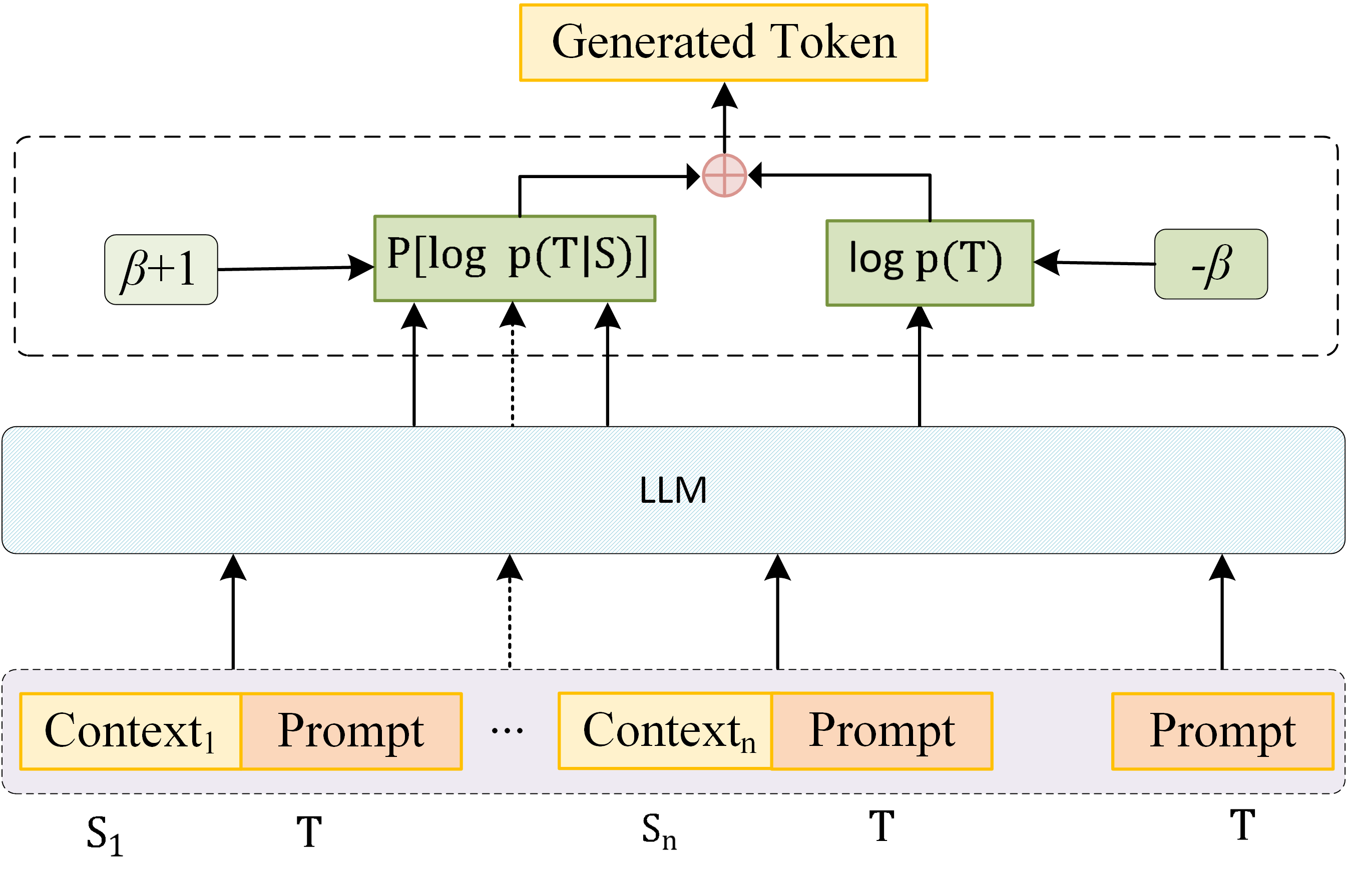}
	\caption{ An example for our NBCE. Initially, NBCE  divides the context into equal-sized windows, each with the maximum length compatible with LLM in-target. Subsequently, a voting mechanism is introduced to select the most relevant context window, regarded as the posterior context. Finally, it employs Bayes' theorem to generate the test task.		
	}
	\label{fig:nbce}
\end{figure}
\begin{equation}
p(S_1,S_2,\ldots,S_n|T) = \prod_{k=1}^{n} p(S_k|T).
\end{equation}
Thus, we have:
\begin{equation}
p(T|S_1,S_2,\ldots,S_n) \propto p(T) \prod_{k=1}^{n} p(S_k|T).
\end{equation}
Furthermore, based on Bayes' formula $p(S_k|T) \propto \frac{p(T|S_k)}{p(T)}$, we get:
\begin{equation}
p(T|S_1,S_2,\ldots,S_n) \propto \frac{1}{{p^{n-1}(T)}} \prod_{k=1}^{n} p(T|S_k).
\end{equation}
Or:

\begin{align}\label{eq:context}
	\log p(T|S_1,S_2,\ldots,S_n) &= \sum_{k=1}^{n} \log p(T|S_k) \nonumber \\
	&\quad - (n-1)\log p(T) \nonumber \\
	&\quad + \text{constant},
\end{align}
where both $p(T|S_k)$ and $p(T)$ can be computed directly utilizing existing LLMs, independent of their architecture, and without the need for fine-tuning on extensive textual data. Specifically, $p(T|S_k)$ represents the probability predicted by an individual contextual set, while $p(T)$ signifies the probability in the absence of any context or with an empty context. It is noteworthy that multiple contextual sets can be concurrently processed within the same batch, with computational complexity scaling linearly with the number of contexts. Certainly, Naive Bayes leans heavily on the independence assumption, which can restrict its practical utility. To aspire to enhance its performance beyond the initial state, we further refine Equation \ref{eq:context}.


%
%

To commence this refinement, we shall introduce the following notations:

\begin{equation}
\log p(T|S) = [\log p(T|S_1), \ldots, \log p(T|S_n)],
\end{equation}
and
\begin{equation}\label{eq:mean}
\overline{\log p(T|S)} = \frac{1}{n} \sum_{k=1}^{n} \log p(T|S_k),
\end{equation}
where $\overline{\log p(T|S)}$ denotes the Average Pooling of $\log p(T|S)$. Let $\beta = n-1$, then Equation \ref{eq:context} can be rewritten as

\begin{align}\label{eq:beta}
	\log p(T|S_1,S_2,\ldots,S_n) &= (\beta+1)\overline{\log p(T|S)} \nonumber \\
	&\quad - \beta\log p(T) \nonumber \\
	&\quad + \text{constant}.
\end{align}



However, the reformulation may prompt the emergence of two inherent inquiries:

\begin{itemize}
\item  If we consider $\beta$ as a hyperparameter subject to tuning, could this potentially yield superior results?
\item  Is it conceivable that employing alternative pooling techniques, denoted as $P$, might potentially yield enhancements in performance? That is:
\end{itemize}

\begin{align}\label{eq:q2}
	\log p(T|S_1,S_2,\ldots,S_n) &= (\beta+1)P[\log p(T|S)] \nonumber \\
	&\quad - \beta\log p(T) \nonumber \\
	&\quad + \text{constant}
\end{align}
To delve deeper into these inquiries, we conducted a series of experiments employing the 7B model and garnered preliminary insights. In the realm of reading comprehension, a consistent trend of robust performance emerges when employing Max Pooling with a $\beta$ value of 0.25 in conjunction with Greedy Search. Conversely, outcomes generated via Random Sampling frequently yield results that are challenging to interpret.

The observed disparities in outcomes can be attributed to the inherent characteristics of these two methods. Random Sampling, characterized by its selection of tokens based on their probability distribution, tends to exhibit lackluster performance, signaling that the output of Max Pooling may not align with a plausible probability distribution.
In contrast, Greedy Search operates distinctively by prioritizing the token with the highest probability, disregarding the holistic distribution. Its commendable performance suggests that the token with the highest probability is more likely to be the accurate choice. Larger probabilities are indicative of lower uncertainty.  To enhance the performance of Random Sampling, we modify the pooling method to directly output the probability distribution  with the lowest uncertainty:

\begin{align}\label{eq:voting}
	P[\log p(T|S)] = \log p(T|S_k),  \nonumber \\
	k = \text{argmin}\{H_1, H_2, \ldots, H_n\}, \nonumber \\
	H_i = -\sum_T p(T|S_i)\log p(T|S_i),
\end{align}

By substituting this expression into Eq.\ref{eq:q2}, we arrive at the conclusive formulation of the NBCE.
It is noteworthy that while the initial inspiration for this approach stemmed from Naive Bayes, the generalized Equation \ref{eq:q2} transcends the conventional boundaries of traditional Naive Bayes, yet maintains its inherent interpretability. Eq.\ref{eq:q2} assumes an intuitive form: Predictions originating from various contextual sources are collectively amalgamated (or weighted) through the utilization of the method denoted as $P$ (with a weight factor of $\beta+1$). Subsequently, this amalgamation is counterbalanced by subtracting the prediction in the absence of context, weighted by $\beta$.
The rationale behind subtracting the context-less prediction lies in enhancing the model's reliance on contextual information, reducing its dependency on inherent knowledge \cite{Shi-2305-14739}. 

The choice of values for $\beta$ can be tailored to different scenarios. For tasks necessitating comprehensive reading comprehension and robust context integration, a larger $\beta$ value may be deemed appropriate. Conversely, tasks leaning towards creative writing may benefit from a smaller $\beta$ value. In our experiments, we set $\beta =0.25$.

\section{Experimental Setup}\label{sec:experimental-setup}
In this section, we describe the experimental settings adopted in our work, including the datasets, LLMs and comparative approaches used to evaluate our approach.

\subsection{Datasets}

In our experiments, we employed a diverse range of benchmark datasets to evaluate our approach. These datasets encompassed various tasks, including text classification and  multiple-choice questions. Fifteen Text Classification Datasets: SST-2 \cite{SocherPWCMNP13}, CR \cite{DingLY08}, RTE \cite{Bar-HaimDS14}, Subj \cite{PangL04}, CB \cite{de2019commitmentbank}, AGNews \cite{ZhangZL15}, SST-5 \cite{SocherPWCMNP13}, YELP \cite{ZhangZL15}, TREC \cite{LiR02}, DBPedia \cite{ZhangZL15}, NLU \cite{LiuESR19}, BANKING77 \cite{abs-2003-04807}, CLINIC150 \cite{LarsonMPCLHKLLT19}, TREC (fine-grained labels) and NLU (fine and coarse-grained labels). Five datasets from Multiple-choice Domain. Specifically, we consider sentence completion: HellaSwag \cite{ZellersHBFC19}; commensense reasoning: PIQA \cite{BiskZLGC20}, OpenBookQA \cite{MihaylovCKS18}, StoryCloze \cite{MostafazadehRLC17}, MMLU \cite{HendrycksBBZMSS21}, ARC-Easy \cite{abs-2102-03315}; and COPA from SuperGLUE benchmark \cite{WangPNSMHLB19}.  It is worth noting that we conducted evaluations using the standard test sets or validation sets when a public test set was not available. It is important to mention that all the datasets used in our experiments are in the English language.

\subsection{Training Sampling and Models}
The effectiveness of ICL has been observed to be highly dependent on the selection of training examples  \cite{ZhaoWFK021}. To ensure a fair and consistent comparison, we maintain the approach employed in the PCW \cite{pcw}, a common practice in prior research \cite{ZhaoWFK021,LuBM0S22,abs-2203-11702}. Specifically, we randomly selected 30 sets from the training datasets and report  the mean and standard deviation calculated across these sampled sets.

Given our limited computational resources, our experiments were conducted using eight large models: GPT2-Large (0.75B), GPT2-XL(1.5B)\cite{radford2019language}, there LLAMA models, including 7B, 13B and 30B \cite{llama_1}, and three OPT models with 1.3B, 6.7B and 13B parameters \cite{abs-2205-01068}.

\subsection{Comparative Baseline}
Note that our proposed solution does not require any additional training. As far as our knowledge extends, \citet{pcw} initiated the work in this line of research. Therefore, we  compare our approach with methods that also do not require further training, as follows.

\begin{itemize}
	\item \textbf{ICL}. A traditional ICL approach employs a conventional single context window, which essentially utilizes the full capacity of the positional embedding in the LLM.
	\item \textbf{PCW}\cite{pcw}. PCW introduces strategic adjustments to both position encoding and attention mask mechanisms to enable  multiple context windows without requiring additional training.
\end{itemize}

\subsection{Prompt Formats}
We have employed the same prompt formats as those adapted by the comparative baseline, PCW. For the sake of brevity, we have omitted specific details about the prompt format; for a more comprehensive understanding, we kindly refer you to \citet{pcw}.

\begin{table*}[t]
	\centering
	\adjustbox{width=\linewidth}{		
		\begin{tabular}{l|c|c|l|ll|ll|ll}  
	\hline					
	\multirow{2}{*}{Dataset}&\multirow{2}{*}{\makecell{\# Shots per\\ window}}&\multirow{2}{*}{\# Labels}&\multirow{2}{*}{\centering \makecell[c]{ICL\\B=1}}&\multicolumn{2}{c|}{B=3}&\multicolumn{2}{c|}{B=6} &\multicolumn{2}{c}{B=9} \\\cline{5-10}
			&&&& \makecell[c]{PCW}& \makecell[c]{NBCE}& \makecell[c]{PCW}& \makecell[c]{NBCE}& \makecell[c]{PCW}& \makecell[c]{NBCE}\\
			\hline

SST-2 		&27&2&  80.2 $\pm$ 11.7 & 84.1 $\pm$ 8.2 & 	\textbf{85.2} $\pm$ 6.7 & 81.2 $\pm$ 7.0 & 	\textbf{83.6} $\pm$ 7.0 & 78.9 $\pm$ 5.3 & 	\textbf{84.3} $\pm$ 5.9$^*$ \\
CR 			&21&2&  81.3 $\pm$ 6.3 & 81.2 $\pm$ 6.4 & 	\textbf{82.7} $\pm$ 6.3 & 82.3 $\pm$ 5.2 & 	\textbf{84.7} $\pm$ 4.6 & 81.2 $\pm$ 3.4 & 	\textbf{84.1} $\pm$ 4.4$^*$ \\
SUBJ 		&18&2&  65.1 $\pm$ 11.9 & 	\textbf{67.0} $\pm$ 12.2 & 66.1 $\pm$ 13.2 & 62.9 $\pm$ 10.9 & 	\textbf{66.2} $\pm$ 10.7 & 60.1 $\pm$ 2.8 & 	\textbf{64.4} $\pm$ 9.9$^*$ \\
CB 			&5&2&  43.9 $\pm$ 3.7 & 43.9 $\pm$ 3.2 & 	\textbf{45.2} $\pm$ 3.7 & 42.8 $\pm$ 2.1 & 	\textbf{44.8} $\pm$ 3.3$^*$ & 42.1 $\pm$ 2.2 & 	\textbf{45.1} $\pm$ 5.0$^*$ \\
RTE 		&5&2&  52.5 $\pm$ 2.2 & 	\textbf{53.5} $\pm$ 1.7 & 52.9 $\pm$ 2.9 &  	\textbf{54.4} $\pm$ 1.0$^*$ & 53.0 $\pm$ 2.4 & 53.9 $\pm$ 2.6 & 	\textbf{54.2} $\pm$ 2.5 \\
AGNews		&11&4&  61.7 $\pm$ 14.2 & 70.9 $\pm$ 9.4 & 	\textbf{71.0} $\pm$ 8.9$^*$ &  	\textbf{67.7} $\pm$ 7.0 & 67.1 $\pm$ 10.6 & 64.8 $\pm$ 3.1 & 	\textbf{72.9} $\pm$ 7.6$^*$ \\
SST5 		&20&5&  40.8 $\pm$ 2.5 & 41.5 $\pm$ 3.1 & 	\textbf{41.8} $\pm$ 2.4 & 37.4 $\pm$ 4.1 & 	\textbf{42.5} $\pm$ 1.9$^*$ & 35.9 $\pm$ 2.8 & 	\textbf{41.9} $\pm$ 2.4$^*$ \\
TREC 		&38&6&  56.6 $\pm$ 7.9 & 59.0 $\pm$ 4.7 & 	\textbf{63.1} $\pm$ 7.0$^*$ & 53.9 $\pm$ 3.1 & 	\textbf{65.3} $\pm$ 3.0$^*$ & 50.9 $\pm$ 3.4 & 	\textbf{66.5} $\pm$ 2.9$^*$ \\
DBPedia 	&7&14&  58.7 $\pm$ 20.2 & 	\textbf{78.9} $\pm$ 6.6$^*$ & 71.1 $\pm$ 13.7 &  	\textbf{79.3} $\pm$ 4.2 & 75.9 $\pm$ 8.2 & 68.1 $\pm$ 1.9 & 	\textbf{76.7} $\pm$ 5.7$^*$ \\
NLU Scenario&43&18&  34.8 $\pm$ 7.6 & 28.5 $\pm$ 4.3 & 	\textbf{45.7} $\pm$ 6.7$^*$ & 26.9 $\pm$ 3.2 & 	\textbf{41.7} $\pm$ 8.5$^*$ & 24.4 $\pm$ 1.6 & 	\textbf{44.1} $\pm$ 6.1$^*$ \\
TREC Fine	&37&50&  31.2 $\pm$ 7.9 & 33.9 $\pm$ 4.4 & 	\textbf{36.9} $\pm$ 6.3$^*$ & 31.3 $\pm$ 3.5 & 	\textbf{40.3} $\pm$ 5.1$^*$ & 26.5 $\pm$ 4.2 & 	\textbf{39.3} $\pm$ 3.9$^*$ \\
NLU Intent 	&43&68&  24.5 $\pm$ 6.1 & 22.3 $\pm$ 5.6 & 	\textbf{27.5} $\pm$ 4.6$^*$ & 19.8 $\pm$ 4.7 & 	\textbf{28.6} $\pm$ 6.1$^*$ & 15.5 $\pm$ 3.4 & 	\textbf{31.1} $\pm$ 4.7$^*$ \\
BANKING77	&27&77&  28.9 $\pm$ 5.1 & 28.0 $\pm$ 3.7 & 	\textbf{36.0} $\pm$ 3.2$^*$ & 23.0 $\pm$ 3.3 & 	\textbf{37.1} $\pm$ 3.4$^*$ & 18.5 $\pm$ 2.7 & 	\textbf{38.5} $\pm$ 3.6$^*$ \\
CLINIC150	&39&150&  43.9 $\pm$ 3.2 & 44.1 $\pm$ 1.9 & 	\textbf{48.5} $\pm$ 2.3$^*$ & 40.4 $\pm$ 1.7 & 	\textbf{49.4} $\pm$ 1.5$^*$ & 35.0 $\pm$ 1.9 & 	\textbf{49.7} $\pm$ 1.8$^*$ \\

							\hline						
		\end{tabular}
	}			
\caption{Comparative Analysis of Classification Accuracy (in \%) for GPT2-Large Using Various Context Windows (B=3, B=6, B=9). \textbf{Note: A single window (B) includes K examples}, falling within the model's capacity (e.g., 1024 tokens in GPT-2). For detailed information on the maximum number of examples (K) for each dataset and model, refer to Appendix Section \ref{sec:prompt}. Best scores are highlighted in bold. An asterisk (*) denotes statistical significance, as determined by a t-test with a p-value < 0.05. The results of GPT-2-Xl are presented in Appendix Table \ref{tab:classification_gpt-xl}. }

	\label{tab:classification_gpt-large}	
\end{table*}

\begin{table*}[t]
	\centering
	\adjustbox{width=\linewidth}{		
		
		\begin{tabular}{l|c|c|l|ll|ll|ll}  
			\hline					
			\multirow{2}{*}{Dataset}&\multirow{2}{*}{\makecell{\# Shots per\\ window}}&\multirow{2}{*}{\# Labels}&\multirow{2}{*}{\makecell[c]{ICL\\B=1}}&\multicolumn{2}{c|}{B=3}&\multicolumn{2}{c|}{B=6} &\multicolumn{2}{c}{B=9} \\\cline{5-10}
			&&&& \makecell[c]{PCW}& \makecell[c]{NBCE}& \makecell[c]{PCW}& \makecell[c]{NBCE}& \makecell[c]{PCW}& \makecell[c]{NBCE}\\
			\hline 

SST-2 		&48&2&  93.4 $\pm$ 1.3 & 	\textbf{94.9} $\pm$ 0.6$^*$ & 93.8 $\pm$ 0.9 & 91.7 $\pm$ 1.0 & 	\textbf{94.0} $\pm$ 0.9$^*$ & 84.5 $\pm$ 0.9 & 	\textbf{94.1} $\pm$ 0.7$^*$ \\
CR 			&39&2&  93.9 $\pm$ 0.7 & 93.5 $\pm$ 0.6 & 	\textbf{94.1} $\pm$ 0.6$^*$ & 90.0 $\pm$ 1.0 & 	\textbf{94.0} $\pm$ 0.5$^*$ & 79.3 $\pm$ 3.3 & 	\textbf{94.2} $\pm$ 0.5$^*$ \\
SUBJ 		&32&2&  70.1 $\pm$ 9.9 & 60.5 $\pm$ 7.6 & 	\textbf{74.2} $\pm$ 7.5$^*$ & 49.8 $\pm$ 1.8 & 	\textbf{69.8} $\pm$ 7.3$^*$ & 48.4 $\pm$ 0.0 & 	\textbf{71.4} $\pm$ 6.9$^*$ \\
CB 			&10&2&  81.3 $\pm$ 5.7 & 	\textbf{81.9} $\pm$ 7.4 & 77.8 $\pm$ 8.3 & 76.4 $\pm$ 5.2 & 	\textbf{78.4} $\pm$ 7.5 & 62.2 $\pm$ 3.0 & 	\textbf{83.9} $\pm$ 3.7$^*$ \\
RTE 		&10&2&  72.9 $\pm$ 3.1 & 	\textbf{73.8} $\pm$ 1.9 & 73.1 $\pm$ 3.1 & 67.2 $\pm$ 2.5 & 	\textbf{74.4} $\pm$ 1.8$^*$ & 57.5 $\pm$ 1.4 & 	\textbf{74.2} $\pm$ 2.4$^*$ \\
AGNews		&20&4&  87.9 $\pm$ 2.8 & 87.3 $\pm$ 1.7 & 	\textbf{88.6} $\pm$ 1.6 & 87.4 $\pm$ 1.1 & 	\textbf{88.8} $\pm$ 1.6$^*$ & 83.1 $\pm$ 1.8 & 	\textbf{89.3} $\pm$ 1.0$^*$ \\
SST5 		&36&5&  40.8 $\pm$ 5.6 & 	\textbf{44.6} $\pm$ 3.8$^*$ & 43.1 $\pm$ 3.5 & 40.4 $\pm$ 4.4 & 	\textbf{42.5} $\pm$ 3.2 & 22.9 $\pm$ 3.0 & 	\textbf{42.9} $\pm$ 2.6$^*$ \\
TREC 		&69&6&  83.4 $\pm$ 5.4 & 81.1 $\pm$ 3.9 & 	\textbf{83.5} $\pm$ 4.7 & 55.1 $\pm$ 3.8 & 	\textbf{86.4} $\pm$ 3.7$^*$ & 41.2 $\pm$ 4.0 & 	\textbf{88.8} $\pm$ 3.0$^*$ \\
DBPedia 	&14&14&  86.7 $\pm$ 6.8 & 	\textbf{94.9} $\pm$ 3.0$^*$ & 93.2 $\pm$ 3.3 &  	\textbf{95.7} $\pm$ 1.6 & 95.6 $\pm$ 2.4 & 92.7 $\pm$ 1.3 & 	\textbf{96.8} $\pm$ 1.3$^*$ \\
NLU Scenario&80&18&  79.6 $\pm$ 3.0 & 79.7 $\pm$ 2.5 & 	\textbf{83.8} $\pm$ 2.2$^*$ & 58.4 $\pm$ 2.9 & 	\textbf{85.0} $\pm$ 1.6$^*$ & 40.4 $\pm$ 4.9 & 	\textbf{86.3} $\pm$ 1.4$^*$ \\
TREC Fine	&65&50&  55.6 $\pm$ 6.1 & 49.5 $\pm$ 5.4 & 	\textbf{57.8} $\pm$ 6.8$^*$ & 33.5 $\pm$ 3.6 & 	\textbf{59.8} $\pm$ 5.0$^*$ & 16.9 $\pm$ 2.9 & 	\textbf{60.9} $\pm$ 4.5$^*$ \\
NLU Intent 	&80&68&  59.9 $\pm$ 5.2 & 	\textbf{62.9} $\pm$ 3.9$^*$ & 54.3 $\pm$ 2.9 & 37.3 $\pm$ 5.6 & 	\textbf{56.6} $\pm$ 3.1$^*$ & 14.8 $\pm$ 3.4 & 	\textbf{57.9} $\pm$ 2.5$^*$ \\
BANKING77	&51&77&  46.3 $\pm$ 4.0 & 	\textbf{51.2} $\pm$ 3.3$^*$ & 50.5 $\pm$ 3.1 & 26.6 $\pm$ 4.5 & 	\textbf{54.6} $\pm$ 3.3$^*$ & 11.2 $\pm$ 3.2 & 	\textbf{58.9} $\pm$ 2.5$^*$ \\
CLINIC150	&72&150&  	\textbf{61.3} $\pm$ 2.5$^*$ & 57.0 $\pm$ 3.2 &  55.4 $\pm$ 2.6 & 32.8 $\pm$ 4.8 & 	\textbf{57.2} $\pm$ 1.8$^*$ & 17.1 $\pm$ 4.0 & 	\textbf{60.8} $\pm$ 1.9$^*$ \\

			\hline				
		\end{tabular}
	}			
	\caption{Comparative Analysis of Classification Accuracy (in \%) for LLAMA-7B Across Various Context Windows. The results of LLAMA-13B and LLAMA-30B  are presented in Appendix Section Tables \ref{tab:classification_llama-13b} and \ref{tab:classification_llama-30b}. }
	\label{tab:classification_llama-7b}	
\end{table*}

\begin{table*}
	\centering
	
	\adjustbox{width=\linewidth}{
		\begin{tabular}{l|c|c|l|ll|ll|ll}  
			\hline					
			\multirow{2}{*}{Dataset}&\multirow{2}{*}{\makecell{\# Shots per\\ window}}&\multirow{2}{*}{\# Labels}&\multirow{2}{*}{\makecell[c]{ICL\\B=1}}&\multicolumn{2}{c|}{B=3}&\multicolumn{2}{c|}{B=6} &\multicolumn{2}{c}{B=9} \\\cline{5-10}
			&&&& \makecell[c]{PCW}& \makecell[c]{NBCE}& \makecell[c]{PCW}& \makecell[c]{NBCE}& \makecell[c]{PCW}& \makecell[c]{NBCE}\\
			\hline

			\hline

			SST-2 			&48&2	& 85.0  $\pm$  8.5 			&  81.7  $\pm$  10.6&	\textbf{86.0  $\pm$  7.2}&81.1  $\pm$  7.7			&  	\textbf{88.1  $\pm$  5.7}* 	& 79.9  $\pm$  9.8          &  	\textbf{88.8  $\pm$  5.2}*  \\
			CR	 			&39&2	& 89.1  $\pm$  2.4 			&  88.8  $\pm$  2.3 &	\textbf{89.7  $\pm$  1.7}&88.5  $\pm$  3.3			&  	\textbf{88.8  $\pm$  1.6} 	& 85.6  $\pm$  3.6      	&  	\textbf{89.1  $\pm$  1.5}* 	  \\
			SUBJ			&32&2	& 	\textbf{78.8  $\pm$  9.0}*&  68.3  $\pm$  7.5 &69.0  $\pm$  7.9&68.5  $\pm$  6.6			& 	\textbf{70.5  $\pm$  7.4 }	& 65.2  $\pm$  8.3       	&  	\textbf{70.9  $\pm$  6.3}* 	 \\
			CB 				&10&2	& 	\textbf{53.0  $\pm$  6.0} &  50.5  $\pm$ 3.3 	&50.8  $\pm$  3.3&	\textbf{51.6  $\pm$  5.2}	&  51.5  $\pm$  4.3 			& 49.1  $\pm$  1.0     		&   \textbf{51.6  $\pm$  3.6}* 	 \\
			RTE				&10&2	&  51.1  $\pm$  3.7 		&  51.8  $\pm$  3.8 &	\textbf{52.7  $\pm$  3.2} &50.6  $\pm$  3.1			&  	\textbf{51.4  $\pm$  2.9 }	& 50.9  $\pm$  2.1          &  	\textbf{51.3  $\pm$  2.5}   \\
			AGNews			&20&4	&  61.3  $\pm$  10.3		& 	\textbf{67.4  $\pm$  6.7}* &59.6  $\pm$  7.2&	\textbf{65.1  $\pm$  5.9}*	&  60.3  $\pm$  9.0 			&	\textbf{ 69.4  $\pm$  5.0}*&  {62.9  $\pm$  6.7} 	 \\
			SST5 			&36&5	& 44.0  $\pm$  3.9 			&  42.7  $\pm$  4.6 &	\textbf{44.8  $\pm$  2.8}&42.4  $\pm$  4.0			& 	\textbf{44.8  $\pm$  2.2}* 	& 41.6  $\pm$  4.3 &         \textbf{45.1  $\pm$  2.0}*   	  \\
			TREC 			&69&6	&	\textbf{59.4  $\pm$  6.3}*&  55.0  $\pm$ 4.3 	&56.8  $\pm$  4.7 &55.2  $\pm$  3.2			&  	\textbf{55.7  $\pm$  4.3} 	& 52.5  $\pm$  2.8          &  	\textbf{57.1  $\pm$  3.9}* 	 \\
			DBPedia			&14&14 &  86.3  $\pm$  3.8 		&  87.7  $\pm$  2.1 &	\textbf{87.9  $\pm$  2.2}&	\textbf{88.1  $\pm$  2.6}	&  87.5  $\pm$  2.6 			& 87.0  $\pm$  3.1          &  	\textbf{87.9  $\pm$  2.6} 	  \\
			NLU Scenario	&80&18	&67.8  $\pm$  4.0 			& 69.9  $\pm$  3.5	&  	\textbf{70.2  $\pm$  4.0}&	\textbf{69.9  $\pm$  2.6}	&  69.3  $\pm$  4.3 			&67.7  $\pm$  4.0           &  	\textbf{72.8  $\pm$  3.8}* \\
			TREC Fine		&65&50	&39.7  $\pm$  4.5 			& 38.8  $\pm$  4.7	& 	\textbf{41.5  $\pm$  6.0}&40.5  $\pm$  5.8			&	\textbf{43.1  $\pm$  6.4} 		&35.3  $\pm$  3.5           &  	\textbf{42.0  $\pm$  4.7}* \\
			NLU Intent		&80&68	&45.3  $\pm$  4.9 			& 50.0  $\pm$  4.2	&  	\textbf{50.9  $\pm$  4.0}&48.8  $\pm$  4.2			& 	\textbf{51.0  $\pm$  4.7} 	&45.4  $\pm$  3.2           &  	\textbf{54.5  $\pm$  3.3}* 	\\
			BANKING77		&51&77	&25.9  $\pm$  4.9 			& 24.8  $\pm$  4.0	&  	\textbf{28.8  $\pm$  4.5} &26.0  $\pm$  3.5			&	\textbf{30.1  $\pm$  3.5}* 	&28.9  $\pm$  3.1           &  	\textbf{32.5  $\pm$  3.5}*  \\
			CLINIC150		&72&150&50.8  $\pm$  3.0 			& 52.4  $\pm$  2.3	& 	\textbf{57.7  $\pm$  2.0}&52.6  $\pm$  2.0			&  	\textbf{57.2  $\pm$  2.5}*	&49.3  $\pm$  2.5           &  	\textbf{58.4  $\pm$  2.0}*	\\
\hline	
	\end{tabular}
	}			
	
	\caption{Comparative Analysis of Classification Accuracy (in \%) for OPT-1.3B models. The results of OPT-6.7B are presented in Appendix Tables \ref{tab:opt_6.7b}. }
	\label{tab:opt_1.3b}	
\end{table*}

\section{Evaluation}
We evaluate our proposed solution based on two primary criteria:
\begin{itemize}
	\item  \textbf{Ability to Extend the Length of Large Models:} Does our solution effectively enable the expansion of the size or capacity of large models?
	\item \textbf{Impact of Additional Demonstrations on ICL Task Performance:} Does the inclusion of more demonstrations have a positive effect on the performance of the ICL task?
\end{itemize}

\subsection{ Classification Task Evaluation}

\subsubsection{Main Results}
\begin{figure}[h]
	\centering
	\includegraphics[width=0.52\textwidth]{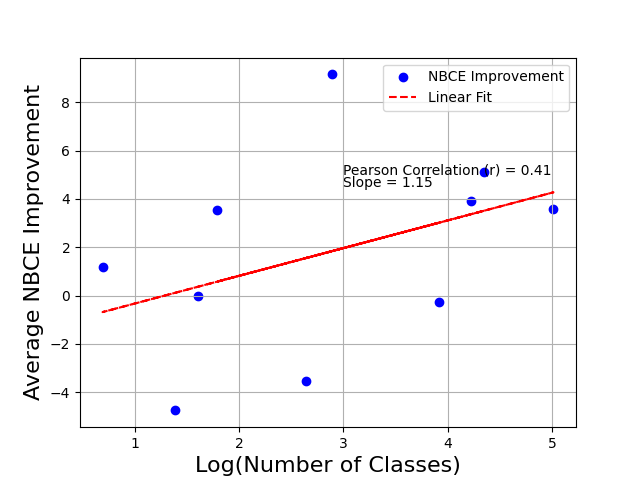}
	\caption{ Average Performance Enhancements with NBCE over PCW as a Function of Label Count: Each data point in our analysis signifies the average improvement observed across all datasets on GPT2 models. It is worth noting a clear and positive correlation between the quantity of unique labels and the benefits derived from our NBCE.}
	\label{fig:correlation}
\end{figure}
We conducted an analysis in which we calculated the average accuracy from 30 different runs, each with a unique seed. We compiled the accuracy and standard deviation for various text classification datasets, which are presented in Tables \ref{tab:classification_gpt-large}, \ref{tab:classification_llama-7b}, and \ref{tab:opt_1.3b}.\textbf{ Due to space constraints, the results of more scaled models are presented in the Appendix Section: GPT2-XL Table \ref{tab:classification_gpt-xl}, LLAMA-13B Table \ref{tab:classification_llama-13b}, LLAMA-30B Table \ref{tab:classification_llama-30b}, and OPT-6.7B Table \ref{tab:opt_6.7b}}. To highlight significant findings, we marked statistical significance with an asterisk (*), based on a t-test with a p-value of less than 0.05. Our key observations are as follows.
(1) Vanilla ICL consistently showed the lowest performance across all models and datasets, underscoring the critical need for expanded context in ICL tasks.
(2) For models with fewer parameters (like GPT-2-Large and OPT-1.3B) and when dealing with a limited number of output classes (five or fewer), we noted minor or negligible differences between both PCW and NBCE, compared to vanilla ICL. Conversely, in models with a larger number of parameters, NBCE generally demonstrated superior performance in most cases. However, it is important to note that several of these differences did not reach statistical significance.
(3) NBCE enhances ICL by accommodating a greater number of examples. This improvement becomes particularly evident when B=9, where both accuracy and stability generally show marked improvements.  We observed that larger models benefit more substantially from our approach. This favorable scaling trend of NBCE is particularly notable when contrasted with previous efforts to enhance ICL (refer to \cite{ZhaoWFK021,LuBM0S22}), where improvements in 178B-scale models were less marked compared to those in smaller models

\subsubsection{PCW enables ICL with a Large Number of Classes}

To investigate the relationship between the number of classes and our NBCE's performance, we conducted a detailed analysis, which was adapted by \citet{pcw}. In each experiment, we calculated the difference between NBCE and PCW and then averaged the results across all datasets on GPT2 models sharing the same number of classes. As illustrated in Figure \ref{fig:correlation}, a robust positive correlation emerged between the quantity of classes and the improvements achieved by NBCE. Specifically, the Pearson correlation coefficient (r) was 0.41 when considering the logarithm of class numbers in relation to the average improvement, with a slope of 1.15. Remarkably, for datasets featuring numerous labels, such as  NLU Intent \cite{LiuESR19}, Banking77 \cite{abs-2003-04807}, and CLINIC150 \cite{LarsonMPCLHKLLT19}, we observed substantial improvements ranging from 3.6 to 5.1 points in most cases. 


When comparing results across datasets with varying numbers of classes, it is crucial to account for potential confounding factors, such as variations in domain, style, or genre. To mitigate these effects, we conducted a comparison using two datasets, each featuring both fine-grained and coarse-grained labels. The TREC dataset \cite{LiR02}, which includes 6 coarse-grained classes. The NLU dataset \cite{LiuESR19}, comprising 18 scenarios coarse-grained classes and 68 intents coarse-grained classes.
Our analysis on GPT2 models, as presented in Table \ref{tab:classification_gpt}, reveals that NBCE outperforms  PCW by  4.1 and 3.0 improvements on GPT2-Large and GPT2-XLarge, respectively. Similarly, in the context of NLU, we observe average improvements  of 17.2 and   5.2 points on GPT2-XLarge, respectively. These findings underscore the effectiveness of our approach, particularly when confronted with a large number of output classes.

\begin{table*}
	\centering
	\adjustbox{width=\linewidth}{		
		\begin{tabular}{l|c|l|ll|ll|ll|ll}  
			\hline					
			\multirow{2}{*}{Dataset}&\multirow{2}{*}{\makecell{\# Shots per\\ window}}&\multirow{2}{*}{\makecell{ICL\\B=1}}&\multicolumn{2}{c|}{B=2}&\multicolumn{2}{c|}{B=3} &\multicolumn{2}{c|}{B=4} &\multicolumn{2}{c}{B=6} \\\cline{4-11}
			&&& \makecell[c]{PCW}& \makecell[c]{NBCE}& \makecell[c]{PCW}& \makecell[c]{NBCE}& \makecell[c]{PCW}& \makecell[c]{NBCE}& \makecell[c]{PCW}& \makecell[c]{NBCE}\\
			
			
			\hline	
			PIQA &23&  81.6 $\pm$ 0.6 & 80.6 $\pm$ 0.7 & 	\textbf{82.1} $\pm$ 0.4$^*$ & 79.6 $\pm$ 0.7 & 	\textbf{82.9} $\pm$ 0.6$^*$ & 79.1 $\pm$ 0.6 & 	\textbf{82.9} $\pm$ 0.6$^*$ & 77.5 $\pm$ 0.8 & 	\textbf{83.0} $\pm$ 0.5$^*$ \\
			OpenBookAQ &63&  41.9 $\pm$ 0.8 & 41.3 $\pm$ 1.0 & 	\textbf{46.3} $\pm$ 0.9$^*$ & 40.9 $\pm$ 0.9 & 	\textbf{49.2} $\pm$ 0.8$^*$ & 39.4 $\pm$ 0.6 & 	\textbf{49.3} $\pm$ 0.9$^*$ & 35.1 $\pm$ 0.8 & 	\textbf{50.3} $\pm$ 1.1$^*$ \\
			COPA &77&  77.8 $\pm$ 1.2 & 	\textbf{78.3} $\pm$ 1.1 & 78.2 $\pm$ 1.5 &  	\textbf{78.9} $\pm$ 1.7$^*$ & 77.5 $\pm$ 1.2 &  	\textbf{77.8} $\pm$ 1.3 & 77.6 $\pm$ 1.6 & 65.9 $\pm$ 3.6 & 	\textbf{76.6} $\pm$ 0.8$^*$ \\
			HellaSwag &12&  79.4 $\pm$ 1.1 & 	\textbf{80.4} $\pm$ 1.1$^*$ & 78.9 $\pm$ 0.9 &  	\textbf{80.2} $\pm$ 0.8$^*$ & 79.6 $\pm$ 0.7 &  	\textbf{80.1} $\pm$ 0.9 & 79.9 $\pm$ 0.8 & 78.5 $\pm$ 0.8 & 	\textbf{79.9} $\pm$ 0.7$^*$ \\
			ARCE &33&  	\textbf{74.4} $\pm$ 1.1$^*$ & 73.8 $\pm$ 1.2 &  72.8 $\pm$ 0.7 &  	\textbf{73.7} $\pm$ 1.4 & 73.5 $\pm$ 0.6 &  	\textbf{74.1} $\pm$ 0.8 & 73.7 $\pm$ 0.8 & 70.8 $\pm$ 1.5 & 	\textbf{73.5} $\pm$ 0.8$^*$ \\
			StoryCloze &24&  46.0 $\pm$ 0.0 & 46.1 $\pm$ 0.1 & 	\textbf{78.7} $\pm$ 0.9$^*$ & 46.1 $\pm$ 0.2 & 	\textbf{78.9} $\pm$ 0.8$^*$ & 46.1 $\pm$ 0.2 & 	\textbf{78.8} $\pm$ 1.0$^*$ & 46.3 $\pm$ 0.2 & 	\textbf{79.6} $\pm$ 0.7$^*$ \\
			MMLU &7&  33.8 $\pm$ 1.9 & 34.1 $\pm$ 2.2 & 	\textbf{34.3} $\pm$ 1.5 & 33.6 $\pm$ 2.3 & 	\textbf{33.7} $\pm$ 1.7 & 34.1 $\pm$ 1.9 & 	\textbf{34.7} $\pm$ 1.9 & 32.5 $\pm$ 3.0 & 	\textbf{33.9} $\pm$ 1.9 \\
				\hline	
			
		\end{tabular}
	}			
	\caption{ Comparative Results of Task Completion (e.g., Multiple Choices Task)  for LLAMA-7B Using  Various Context Windows. Best scores are highlighted in bold. An asterisk (*) denotes statistical significance, as determined by a t-test with a p-value < 0.05. The results of LLAMA-13B are presented in Appendix Tables \ref{tab:multi_choice_llama_13b}. }
	\label{tab:multi_choice_llama_7b}	
\end{table*}

\begin{table*}
	\centering
	
	\adjustbox{width=\linewidth}{		
		\begin{tabular}{l|c|ll|ll|ll|ll}  
			\hline					
			\multirow{2}{*}{Dataset}&\multirow{2}{*}{\# Labels}&\multicolumn{2}{c|}{GPT2-Large}&\multicolumn{2}{c|}{GPT2-XL} &\multicolumn{2}{c|}{LLAMA-7B}&\multicolumn{2}{c}{LLAMA-13B} \\\cline{3-10}
			&& \makecell[c]{NBCE (RAN)}& \makecell[c]{NBCE}&\makecell[c]{NBCE (RAN)}& \makecell[c]{NBCE}&\makecell[c]{NBCE (RAN)}& \makecell[c]{NBCE}&\makecell[c]{NBCE (RAN)}& \makecell[c]{NBCE}\\
			\hline 

			SST-2 		&2&  80.5 $\pm$ 4.5 & \textbf{84.3} $\pm$ 5.9$^*$ 	& 91.6 $\pm$ 1.5 & \textbf{92.5} $\pm$ 1.5 	& 92.3 $\pm$ 1.5 & \textbf{94.1} $\pm$ 0.7$^*$ 	& 92.2 $\pm$ 1.0 & \textbf{94.9} $\pm$ 0.5$^*$ 	\\
			CR 			&2&  78.0 $\pm$ 3.9 & \textbf{84.1} $\pm$ 4.4$^*$ 	& 81.0 $\pm$ 2.2 & \textbf{81.9} $\pm$ 2.0 	& 91.9 $\pm$ 1.2 & \textbf{94.2} $\pm$ 0.5$^*$ 	& 91.1 $\pm$ 1.3 & \textbf{93.1} $\pm$ 0.6$^*$ 	\\
			SUBJ 		&2&  57.0 $\pm$ 3.8 & \textbf{64.4} $\pm$ 9.9$^*$ 	& 72.0 $\pm$ 5.0 & \textbf{76.0} $\pm$ 7.0 	& 69.0 $\pm$ 3.4 & \textbf{71.4} $\pm$ 6.9 	& 89.9 $\pm$ 3.0 & \textbf{93.0} $\pm$ 1.7$^*$ 	\\
			CB 			&2& \textbf{46.1} $\pm$ 4.4 & 45.1 $\pm$ 5.0 	& \textbf{55.3} $\pm$ 6.2 & 54.8 $\pm$ 8.5 	&  81.6 $\pm$ 5.1 & \textbf{83.9} $\pm$ 3.7$^*$ 	& 81.7 $\pm$ 4.0 & \textbf{84.1} $\pm$ 3.5$^*$ 	\\
			RTE 		&2&  52.5 $\pm$ 2.8 & \textbf{54.2} $\pm$ 2.5 	& 53.9 $\pm$ 2.9 & \textbf{55.3} $\pm$ 2.2 	& 68.2 $\pm$ 1.9 & \textbf{74.2} $\pm$ 2.4$^*$ 	& 72.9 $\pm$ 2.3 & \textbf{75.1} $\pm$ 1.5$^*$ 	\\
			AGNews		&4&  66.4 $\pm$ 7.5 & \textbf{72.9} $\pm$ 7.6$^*$ 	& 69.5 $\pm$ 5.9 & \textbf{76.3} $\pm$ 4.7$^*$ 	& 83.4 $\pm$ 2.1 & \textbf{89.3} $\pm$ 1.0$^*$ 	& 85.3 $\pm$ 2.3 & \textbf{87.9} $\pm$ 1.1$^*$ 	\\
			SST5 		&5&  41.3 $\pm$ 1.8 & \textbf{41.9} $\pm$ 2.4 	& 39.1 $\pm$ 3.6 & \textbf{41.7} $\pm$ 5.3 	& 40.4 $\pm$ 2.7 & \textbf{42.9} $\pm$ 2.6$^*$ 	& 44.5 $\pm$ 2.1 & \textbf{47.7} $\pm$ 2.0$^*$ 	\\
			TREC 		&6&  61.0 $\pm$ 2.8 & \textbf{66.5} $\pm$ 2.9$^*$ 	& 50.7 $\pm$ 2.8 & \textbf{51.6} $\pm$ 3.0 	& 84.1 $\pm$ 3.5 & \textbf{88.8} $\pm$ 3.0$^*$ 	& 81.7 $\pm$ 4.4 & \textbf{85.0} $\pm$ 2.4$^*$ 	\\
			DBPedia 	&14&  68.9 $\pm$ 8.2 & \textbf{76.7} $\pm$ 5.7$^*$ 	& 84.1 $\pm$ 2.5 & \textbf{89.0} $\pm$ 2.8$^*$ 	& 82.8 $\pm$ 2.7 & \textbf{96.8} $\pm$ 1.3$^*$ 	& 89.2 $\pm$ 3.4 & \textbf{96.9} $\pm$ 1.3$^*$ 	\\
			NLU Scenario&18&  40.8 $\pm$ 4.8 & \textbf{44.1} $\pm$ 6.1 	& 45.3 $\pm$ 3.9 & \textbf{55.1} $\pm$ 5.4$^*$ 	& 82.0 $\pm$ 2.1 & \textbf{86.3} $\pm$ 1.4$^*$ 	& 81.7 $\pm$ 1.8 & \textbf{88.7} $\pm$ 1.0$^*$ 	\\
			TREC Fine	&50&  33.2 $\pm$ 4.2 & \textbf{39.3} $\pm$ 3.9$^*$ 	& 35.2 $\pm$ 4.4 & \textbf{41.9} $\pm$ 3.7$^*$ 	& 56.7 $\pm$ 3.1 & \textbf{60.9} $\pm$ 4.5$^*$ 	& 57.1 $\pm$ 3.5 & \textbf{63.3} $\pm$ 4.1$^*$ 	\\
			NLU Intent 	&68&  28.3 $\pm$ 0.8 & \textbf{31.1} $\pm$ 4.7$^*$ 	& 35.1 $\pm$ 1.2 & \textbf{40.3} $\pm$ 3.6$^*$ 	& 57.2 $\pm$ 2.1 & \textbf{57.9} $\pm$ 2.5$^*$ 	&  \textbf{62.6} $\pm$ 2.4$^*$ &61.8 $\pm$ 2.1 	\\
			BANKING77	&77&  29.3 $\pm$ 1.6 & \textbf{38.5} $\pm$ 3.6$^*$ 	& 33.6 $\pm$ 1.3 & \textbf{38.9} $\pm$ 2.4$^*$ 	& 47.0 $\pm$ 1.5 & \textbf{58.9} $\pm$ 2.5$^*$ 	& 48.7 $\pm$ 3.2 & \textbf{63.5} $\pm$ 2.3$^*$ 	\\
			CLINIC150	&150&  43.8 $\pm$ 1.7 & \textbf{49.7} $\pm$ 1.8$^*$ 	& 47.7 $\pm$ 1.1 & \textbf{51.6} $\pm$ 1.7$^*$ 	& 58.7 $\pm$ 2.1 & \textbf{60.8} $\pm$ 1.9$^*$ 	& 62.5 $\pm$ 2.2 & \textbf{66.2} $\pm$ 2.2$^*$ 	\\
			\hline						
		\end{tabular}
	}			
	\caption{ Ablation Study with Context Window B=9. Best scores are highlighted in bold. An asterisk (*) denotes statistical significance, as determined by a t-test with a p-value < 0.05. 
	}
	\label{tab:classification_abil_llama-7b}	
\end{table*}

\subsection{Multi-Choice Tasks}
Table \ref{tab:multi_choice_llama_7b} shows the evaluation of multi-choice tasks. It is important to note that the improvements made by both PCW and our NBCE in these tasks, compared to text classification, are relatively modest, with a slight edge for NBCE. Furthermore, employing a greater number of demonstrations does not consistently translate to better performance in multi-choice tasks. Instead, we observe  that scaling up the model size (Appendix Section Table \ref{tab:multi_choice_llama_13b}), rather than increasing the number of demonstrations, tends to yield more substantial improvements in these tasks.

\subsection{Impact of more Demonstrations on ICL}

We conducted experiments to validate the impact of additional demonstrations on ICL in NLP models. Our focus was to show how extra demonstrations (B=6 and B=9, where B is the window size) enhance model performance by improving context understanding and robustness. \textbf{Note that each window contains $K$ samples within the model's token limit (e.g., 2024 tokens for LLAMA). } For detailed information on the maximum value of K for each model and dataset, please see Appendix Table \ref{tab:format} . This approach aligns with the importance of training example quantity in model adaptability and generalization \cite{tacl_a_00574,AhmedWAPSCL24}.
 Our observations indicate that NBCE mostly outperforms its counterpart, PCW, and these improvements can be considered significant. Additionally, scaling up the model size (Appendix Section Tables \ref{tab:classification_gpt-xl},\ref{tab:classification_llama-13b}, \ref{tab:classification_llama-30b}, and \ref{tab:opt_6.7b}) leads to improved performance, especially on larger and more complex datasets.

\subsection{Ablation Study}

To better evaluate the proposed voting mechanism, i.e., selecting the best \( k \) contexts as the posterior in Equation \ref{eq:voting}, we conducted an ablation study introducing a new variant, referred to as NBCE (RAND). In this variant, rather than deliberately choosing \( k \), we randomly select one context from the context windows. The results are presented in Table \ref{tab:classification_abil_llama-7b}. The experimental outcomes across a variety of models and datasets demonstrate that a careful selection of \( k \) significantly contributes to the quality of the generated tokens. It is noteworthy that, in this setting, NBCE can be considered as a standard ICL, where only one context window is considered. However, the performance may slightly differ due to the likelihood of the generated text \( p(T) \), as outlined in Equation \ref{eq:q2}, affecting the final performance.

\subsection{Effect of Pooling Mechanism}

\begin{figure}[h]
	\centering
	\includegraphics[width=0.48\textwidth]{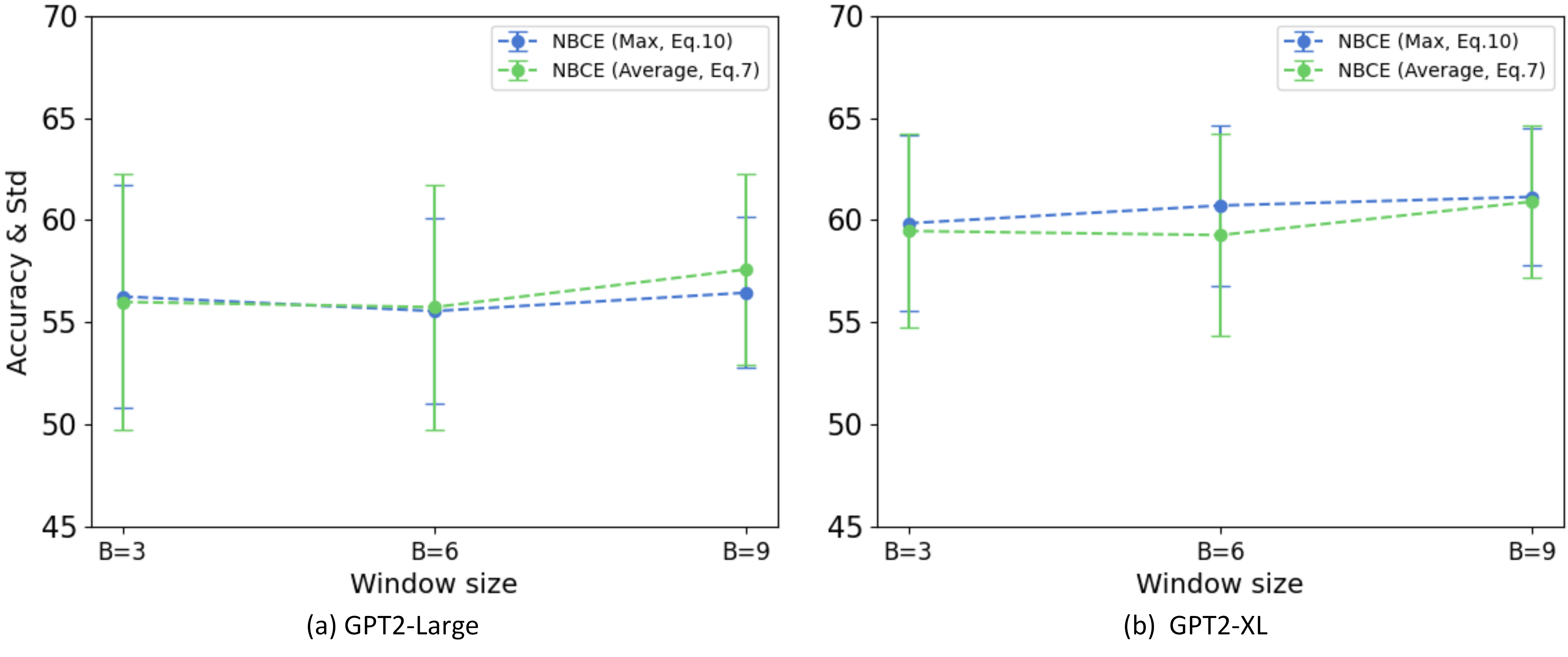}
	\caption{Efficacy in terms of averaged accuracy and standard deviation (i.e., the error bars) of two pooling mechanisms: average context window (Eq.\ref{eq:mean}) and entropy-based maximization (Eq.\ref{eq:voting}) utilizing GPT2 models for text classification. Notably, the maximizing approach enhances both accuracy and stability, with model size impacting averaging pooling's performance.}
	\label{fig:mean}
\end{figure}

It is imperative to underscore the introduction of two distinct pooling mechanisms \(P\): averaging the context windows as depicted in Eq.\ref{eq:mean}, and maximizing based on entropy as delineated in Eq.\ref{eq:voting}. To empirically validate the efficacy of these pooling strategies, we conducted a series of experiments utilizing GPT2 models within the context of a text classification task. The outcomes, illustrated in Figure \ref{fig:mean}, showcase a comparative analysis in terms of both accuracy and standard deviation. Notably, the maximizing strategy not only augments performance but also enhances stability. It is important to acknowledge that the model size can significantly influence the outcomes when employing the averaging pooling mechanism.

\subsection{Effect of \(\beta\)}
\begin{figure}[h]
	\centering
	\includegraphics[width=0.48\textwidth]{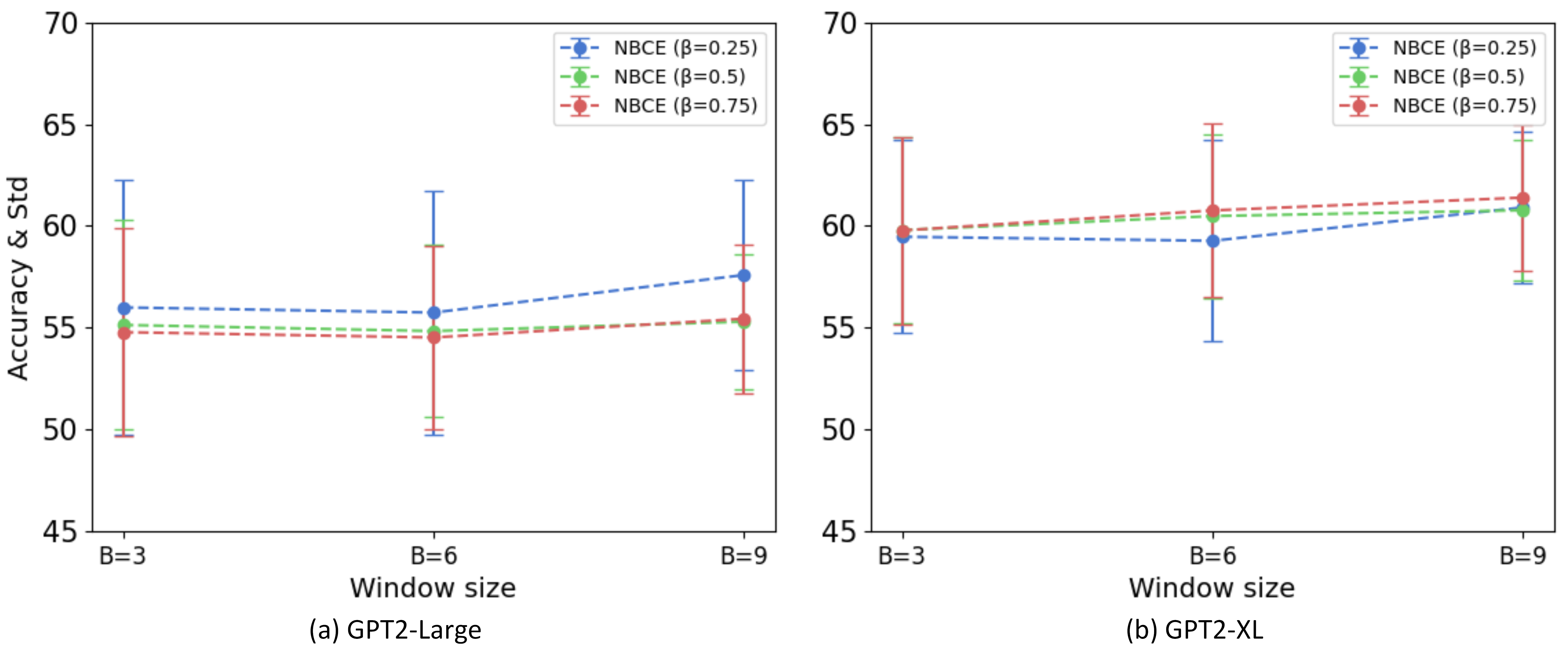}
	\caption{Comparative analysis in terms of averaged accuracy and standard deviation	(i.e., the error bars)  of GPT2 model performance across varying \(\beta\)  Eq.~\ref{eq:beta} values (0.25, 0.5, 0.75) in a text classification task.}

	\label{fig:beta}
\end{figure}

In our investigation, the parameter \(\beta\) as outlined in Eq.~\ref{eq:beta} was initially set to 0.25. To elucidate the impact of different \(\beta\) values on the overall performance, experiments were conducted utilizing GPT2 models within a text classification framework, testing \(\beta\) values of 0.25, 0.5, and 0.75. The outcomes, depicted in Figure~\ref{fig:beta}, illustrate comparative performances in terms of accuracy and standard deviation. The analysis reveals a reduced sensitivity of model performance to variations in \(\beta\), indicating that modifications to \(\beta\) do not markedly influence model robustness. Notably, an observation was made that larger model sizes exhibit more stable performance at increased \(\beta\) values, particularly at \(\beta=0.75\). This stability accentuates the capability of larger models to manage greater parameter variability, enhancing their utility in a broad spectrum of computational tasks.

\section{Related Work}\label{sec:related_work}

%
%
%

\subsection{In-Context Learning}

In recent years, in-context learning has received significant attention in the research community. \citet{BrownMRSKDNSSAA20} introduced this concept, sparking a wave of investigations. \citet{ZhaoWFK021,HanH0SW23} addressed the issue of LLM miscalibrations and explored various calibration methods. 
However, few-shot performance can vary based on the order of demonstrations and template choices \cite{LuBM0S22}. In this context, \citet{ZhaoWFK021} identified three biases and suggested content-free output calibration. \citet{MinLHZ22} demonstrated how these biases shift decision boundaries and proposed calibrating through prototypical cluster distribution estimation. Others focused on prompt engineering, such as selecting optimal demonstration permutations \cite{LuBM0S22}  and using retrieval modules for semantically similar in-context examples \cite{LiuSZDCC22,RubinHB22}. One promising direction is to improve in-context learning by increasing the number of demonstrations.

\subsection{Context Extension}
Expanding the contextual capabilities of LLM continues to pose a formidable challenge and has attracted considerable research attention. Various studies have introduced to tackle the memory limitations associated with self-attention mechanisms. These approaches can be broadly classified into two categories: fine-tuned approaches and few-shot approaches. 
\citet{ZaheerGDAAOPRWY20,GuoAUONSY22}, have suggested using sparse attention as a solution to this issue. \citet{PressSL22} took a novel approach by incorporating positional information using relative factors in attention weights instead of relying on absolute positional encoding. Despite the impressive capabilities of \citet{PressSL22}'s model for extrapolation, it remains computationally intensive due to its quadratic self-attention cost, making it slow and resource-demanding for longer prompts. \citet{abs-2208-00748}  introduced an alternative approach called SLED, which is an encoder-decoder model specifically designed for handling lengthy texts. This model encodes short overlapping segments of input text and integrates this information within the decoder, similar to the Fusion-in-Decoder concept by \citet{IzacardG21}. However, these researches require additional training.

More recently, \citet{pcw} have introduced the concept of Parallel Context Windows (PCW), which enables the concurrent utilization of multiple context windows without requiring additional training. PCW has been purposefully tailored for self-attention models, involving modifications to both position encoding and attention mask mechanisms to enhance the performance.
NBCE and PCW share noteworthy similarities, as they both treat contexts as unordered and apply equal weighting. Notably, when NBCE is employed within the context of a single-layer, single-head attention model, the resulting outcomes closely approximate those achieved through the utilization of PCW. To substantiate this claim, we can formulate the language model tailored to a single-layer, single-head attention configuration.

\begin{equation}
	p(x_t|x_{<t}) = \text{softmax}\left(\sum_{i=1}^{t} a_{t,i}  v_i  W \right)
\end{equation}
hence, approximately: $	\log p(x_t|x_{<t}) \sim \sum_{i=1}^{t} a_{t,i}  v_i  W$. Substituting this into Equation \ref{eq:q2} and setting $\beta = 0$, we obtain:

\begin{align}
	\log p(T|S_1, S_2, \ldots, S_n) &\sim \frac{1}{n} \sum_{k=1}^{n} \left(\sum_{i \in S_k} a_{T,i} v_i \right) W \nonumber \\
	&= \left(\sum_{i \in S_1 \oplus \ldots \oplus S_n} \frac{ a_{T,i}}{n}   v_i\right) W
\end{align}
here, we assume $T$ represents a single sequence (i.e., the query), However, this assumption does not lack generality. The symbol $\oplus$ denotes concatenation and $S_k \oplus T$ is used for reasoning as a continuous segment (as per NBCE's setup), so their positional encodings are adjacent. Additionally, $a_{T,i}/n$ forms a collective attention for  $T$  with all $S_i$ (with a sum equal to 1). These characteristics are consistent with PCW, which is essentially integrated into each layer more elegantly through an attention mask. Therefore, PCW can be thought of as a version of NBCE that utilizes average pooling.


\section{Conclusion}\label{sec:conclusion}
This paper introduces a novel framework called Naive Bayes-based Context Extension (NBCE) for large language models. NBCE innovatively incorporates a voting mechanism to select the most appropriate window context, and then utilizes Bayes' theorem to generate the task text. Our results show  that NBCE outperforms its alternative PCW across a diverse set of multi-class classification tasks. For future work, while PCW shows effective without additional training, ICL could potentially benefit from more demonstrations in fine-tuning settings; however, further investigation is required  to fully comprehend the extent of its advantages. 


\section*{Limitations}

NBCE facilitates ICL tasks by allowing for more demonstrations without the need for fine-tuning. However, there are still some limitations to this approach:

\begin{itemize}
	\item Since NBCE essentially functions as a voting mechanism, its effectiveness is constrained in tasks that require ordered or interrelated contexts, such as code generation. This is due to its inherent nature, which may not adequately handle sequential or dependent information in certain contexts.
	\item Increasing the number of shots does not necessarily lead to improved performance. Experimental results have indicated that expanding the context window size does not significantly enhance performance in completion tasks. This suggests a diminishing return on performance gains with an increased number of contexts.
\end{itemize}



\bibliography{anthology,custom}

\appendix

\section{Appendix}\label{sec:appendix}

\subsection{Scaling Model Parameters}  \label{sec:scaling}

\begin{table*}[t]
	\centering
	\adjustbox{width=\linewidth}{		
		\begin{tabular}{l|c|c|ll|ll|ll}  
			\hline					
			\multirow{2}{*}{Dataset}&\multirow{2}{*}{\# Labels}&\multirow{2}{*}{ICL}&\multicolumn{2}{c|}{B=3}&\multicolumn{2}{c|}{B=6} &\multicolumn{2}{c}{B=9} \\\cline{4-9}
			&&& \makecell[c]{PCW}& \makecell[c]{NBCE}& \makecell[c]{PCW}& \makecell[c]{NBCE}& \makecell[c]{PCW}& \makecell[c]{NBCE}\\

			\hline		
			
			SST-2 		&2&  90.6 $\pm$ 3.5 & 92.4 $\pm$ 2.5 & \textbf{92.7} $\pm$ 2.3$^*$ & 89.4 $\pm$ 3.5 & \textbf{92.5} $\pm$ 2.2$^*$ & 83.7 $\pm$ 1.7 & \textbf{92.5} $\pm$ 1.5$^*$ \\
			CR 			&2&  79.2 $\pm$ 5.9 & 81.3 $\pm$ 4.6 & \textbf{82.5} $\pm$ 2.9$^*$ & 81.6 $\pm$ 2.4 & \textbf{81.9} $\pm$ 2.1 &  \textbf{82.7} $\pm$ 1.7 & 81.9 $\pm$ 2.0 \\
			SUBJ 		&2&  68.8 $\pm$ 11.6 & 64.9 $\pm$ 7.3 & \textbf{74.5} $\pm$ 8.3$^*$ & 57.0 $\pm$ 4.1 & \textbf{78.7} $\pm$ 4.8$^*$ & 65.6 $\pm$ 3.0 & \textbf{76.0} $\pm$ 7.0$^*$ \\
			CB 			&2&  51.9 $\pm$ 7.4 & \textbf{57.2} $\pm$ 8.5$^*$ & 56.1 $\pm$ 7.9 & 49.6 $\pm$ 3.6 & \textbf{55.8} $\pm$ 7.8$^*$ & 42.2 $\pm$ 2.1 & \textbf{54.8} $\pm$ 8.5$^*$ \\
			RTE 		&2&  55.4 $\pm$ 2.4 & \textbf{55.6} $\pm$ 1.6 & 54.9 $\pm$ 2.5 & 54.2 $\pm$ 1.3 & \textbf{55.2} $\pm$ 2.3$^*$ & 50.4 $\pm$ 2.0 & \textbf{55.3} $\pm$ 2.2$^*$ \\
			AGNews		&4&  67.2 $\pm$ 13.2 & \textbf{79.6} $\pm$ 3.4$^*$ & 70.0 $\pm$ 9.6 &  \textbf{80.4} $\pm$ 2.3$^*$ & 74.1 $\pm$ 5.8 & 71.6 $\pm$ 2.5 & \textbf{76.3} $\pm$ 4.7$^*$ \\
			SST5 		&5&  38.0 $\pm$ 6.1 & \textbf{41.4} $\pm$ 4.3$^*$ & 41.1 $\pm$ 4.7 & 38.1 $\pm$ 3.6 & \textbf{41.5} $\pm$ 5.4$^*$ & 35.3 $\pm$ 2.2 & \textbf{41.7} $\pm$ 5.3$^*$ \\
			TREC 		&6&  47.9 $\pm$ 5.1 & 48.7 $\pm$ 2.8 & \textbf{51.7} $\pm$ 5.0$^*$ & 45.5 $\pm$ 2.3 & \textbf{51.8} $\pm$ 4.6$^*$ & 43.1 $\pm$ 1.9 & \textbf{51.6} $\pm$ 3.0$^*$ \\
			DBPedia 	&14&  77.5 $\pm$ 9.8 & 87.0 $\pm$ 4.0 & \textbf{87.7} $\pm$ 3.8$^*$ &  \textbf{88.9} $\pm$ 3.3 & 88.6 $\pm$ 3.3 & 81.4 $\pm$ 2.1 & \textbf{89.0} $\pm$ 2.8$^*$ \\
			NLU Scenario&18&  45.1 $\pm$ 9.3 & 50.0 $\pm$ 6.1 & \textbf{51.1} $\pm$ 8.1$^*$ & 46.7 $\pm$ 5.9 & \textbf{50.3} $\pm$ 6.8$^*$ & 38.7 $\pm$ 6.3 & \textbf{55.1} $\pm$ 5.4$^*$ \\
			TREC Fine	&50&  36.4 $\pm$ 6.2 & 40.0 $\pm$ 3.0 & \textbf{40.1} $\pm$ 5.1$^*$ & 35.5 $\pm$ 2.6 & \textbf{41.7} $\pm$ 3.6$^*$ & 31.0 $\pm$ 2.8 & \textbf{41.9} $\pm$ 3.7$^*$ \\
			NLU Intent 	&68&  30.2 $\pm$ 5.4 & 33.8 $\pm$ 4.6 & \textbf{36.4} $\pm$ 4.9$^*$ & 33.4 $\pm$ 4.3 & \textbf{38.5} $\pm$ 5.4$^*$ & 24.3 $\pm$ 3.7 & \textbf{40.3} $\pm$ 3.6$^*$ \\
			BANKING77	&77&  30.7 $\pm$ 4.1 & 33.3 $\pm$ 3.5 & \textbf{35.5} $\pm$ 2.8$^*$ & 26.8 $\pm$ 3.1 & \textbf{37.6} $\pm$ 2.4$^*$ & 16.7 $\pm$ 2.6 & \textbf{38.9} $\pm$ 2.4$^*$ \\
			CLINIC150	&150&  46.6 $\pm$ 2.5 & 47.1 $\pm$ 2.3 & \textbf{49.9} $\pm$ 1.9$^*$ & 40.8 $\pm$ 2.3 & \textbf{50.9} $\pm$ 2.1$^*$ & 34.5 $\pm$ 2.5 & \textbf{51.6} $\pm$ 1.7$^*$ \\
			\hline

		\end{tabular}
	}			
	\caption{  Comparative Analysis of Classification Accuracy (in \%) for GPT-2-XL Across Various Context Windows (B=3, B=6, B=9). Best scores are highlighted in bold. An asterisk (*) denotes statistical significance, as determined by a t-test with a p-value < 0.05. }
	\label{tab:classification_gpt-xl}	
\end{table*}

\begin{table*}[t]
	\centering
	\adjustbox{width=\linewidth}{		
		\begin{tabular}{l|c|c|ll|ll|ll}  
			\hline					
			\multirow{2}{*}{Dataset}&\multirow{2}{*}{\# Labels}&\multirow{2}{*}{ICL}&\multicolumn{2}{c|}{B=3}&\multicolumn{2}{c|}{B=6} &\multicolumn{2}{c}{B=9} \\\cline{4-9}
			&&& \makecell[c]{PCW}& \makecell[c]{NBCE}& \makecell[c]{PCW}& \makecell[c]{NBCE}& \makecell[c]{PCW}& \makecell[c]{NBCE}\\
			\hline

			SST-2 		&2&  94.5 $\pm$ 0.7 & 94.1 $\pm$ 0.7 & \textbf{94.8} $\pm$ 0.5$^*$ & 94.0 $\pm$ 0.9 & \textbf{95.0} $\pm$ 0.4$^*$ & 90.1 $\pm$ 1.2 & \textbf{94.9} $\pm$ 0.5$^*$ \\
			CR 			&2&  92.0 $\pm$ 1.4 & 92.2 $\pm$ 0.9 & \textbf{92.9} $\pm$ 1.0$^*$ & 92.5 $\pm$ 0.5 & \textbf{93.0} $\pm$ 1.0$^*$ & 91.1 $\pm$ 0.9 & \textbf{93.1} $\pm$ 0.6$^*$ \\
			SUBJ 		&2&  90.2 $\pm$ 3.8 & 87.5 $\pm$ 3.3 & \textbf{90.8} $\pm$ 2.9$^*$ & 79.0 $\pm$ 7.2 & \textbf{92.5} $\pm$ 1.7$^*$ & 67.1 $\pm$ 5.4 & \textbf{93.0} $\pm$ 1.7$^*$ \\
			CB 			&2&  80.3 $\pm$ 8.0 & \textbf{84.6} $\pm$ 4.1$^*$ & 79.8 $\pm$ 4.9 &  \textbf{83.1} $\pm$ 4.0$^*$ & 80.3 $\pm$ 6.4 & 74.1 $\pm$ 6.3 & \textbf{84.1} $\pm$ 3.5$^*$ \\
			RTE 		&2&  \textbf{74.6} $\pm$ 2.7 & 73.5 $\pm$ 2.0 &  74.0 $\pm$ 2.5 & 71.9 $\pm$ 1.6 & \textbf{74.6} $\pm$ 1.6$^*$ & 66.4 $\pm$ 2.0 & \textbf{75.1} $\pm$ 1.5$^*$ \\
			AGNews		&4&  86.9 $\pm$ 2.9 & \textbf{87.9} $\pm$ 1.7 & 86.6 $\pm$ 1.8 &  \textbf{88.0} $\pm$ 0.9 & 87.3 $\pm$ 1.8 & 87.7 $\pm$ 1.1 & \textbf{87.9} $\pm$ 1.1 \\
			SST5 		&5&  48.0 $\pm$ 3.3 & \textbf{49.2} $\pm$ 2.6 & 48.0 $\pm$ 3.3 &  \textbf{48.4} $\pm$ 2.1 & 47.3 $\pm$ 3.4 & 44.0 $\pm$ 2.9 & \textbf{47.7} $\pm$ 2.0$^*$ \\
			TREC 		&6&  83.1 $\pm$ 3.1 & \textbf{83.7} $\pm$ 2.9$^*$ & 81.5 $\pm$ 3.4 & 75.5 $\pm$ 3.6 & \textbf{83.0} $\pm$ 3.8$^*$ & 49.5 $\pm$ 5.4 & \textbf{85.0} $\pm$ 2.4$^*$ \\
			DBPedia 	&14&  88.6 $\pm$ 6.1 & \textbf{93.6} $\pm$ 3.9$^*$ & 93.2 $\pm$ 3.9 & 94.4 $\pm$ 2.7 & \textbf{94.7} $\pm$ 2.6 & 94.5 $\pm$ 2.7 & \textbf{96.9} $\pm$ 1.3$^*$ \\
			NLU Scenario&18&  82.1 $\pm$ 2.7 & 85.9 $\pm$ 1.8 & \textbf{86.7} $\pm$ 1.8$^*$ & 81.2 $\pm$ 2.4 & \textbf{87.4} $\pm$ 1.4$^*$ & 74.1 $\pm$ 2.9 & \textbf{88.7} $\pm$ 1.0$^*$ \\
			TREC Fine	&50&  55.4 $\pm$ 5.3 & \textbf{60.1} $\pm$ 5.1$^*$ & 57.7 $\pm$ 4.7 & 56.8 $\pm$ 5.4 & \textbf{60.4} $\pm$ 4.7$^*$ & 47.6 $\pm$ 9.0 & \textbf{63.3} $\pm$ 4.1$^*$ \\
			NLU Intent 	&68&  68.3 $\pm$ 4.1 & \textbf{73.0} $\pm$ 2.6$^*$ & 58.1 $\pm$ 2.3 &  \textbf{65.2} $\pm$ 2.6$^*$ & 60.7 $\pm$ 2.7 & 52.6 $\pm$ 3.6 & \textbf{61.8} $\pm$ 2.1$^*$ \\
			BANKING77	&77&  46.6 $\pm$ 4.2 & \textbf{56.4} $\pm$ 2.8$^*$ & 52.8 $\pm$ 3.5 & 50.8 $\pm$ 3.1 & \textbf{59.2} $\pm$ 2.8$^*$ & 40.2 $\pm$ 2.5 & \textbf{63.5} $\pm$ 2.3$^*$ \\
			CLINIC150	&150&  63.7 $\pm$ 2.5 & \textbf{66.0} $\pm$ 2.7$^*$ & 59.2 $\pm$ 2.3 & 57.5 $\pm$ 2.9 & \textbf{62.4} $\pm$ 1.7$^*$ & 48.7 $\pm$ 2.3 & \textbf{66.2} $\pm$ 2.2$^*$ \\

			\hline			
		\end{tabular}
	}			
	\caption{Comparative Analysis of Classification Accuracy (in \%) for LLAMA-13B Across Various Context Windows (B=3, B=6, B=9). Best scores are highlighted in bold. An asterisk (*) denotes statistical significance, as determined by a t-test with a p-value < 0.05.  }
	\label{tab:classification_llama-13b}	
\end{table*}

\begin{table*}[t]
	\centering
		\begin{tabular}{l|c|c|ll|ll}  
			\hline					
			\multirow{2}{*}{Dataset}&\multirow{2}{*}{\# Labels}&\multirow{2}{*}{ICL}&\multicolumn{2}{c|}{B=3}&\multicolumn{2}{c}{B=6}  \\\cline{4-7}
			&&& \makecell[c]{PCW}& \makecell[c]{NBCE}& \makecell[c]{PCW}& \makecell[c]{NBCE}\\
			
			\hline			
			
			SST-2 		&2&  94.7 $\pm$ 0.5 & 94.9 $\pm$ 0.7 & \textbf{95.0} $\pm$ 0.3 & 92.9 $\pm$ 0.7 & \textbf{95.0} $\pm$ 0.3$^*$ \\
			CR 			&2&  \textbf{93.8} $\pm$ 0.5 & 93.6 $\pm$ 0.5 &  93.8 $\pm$ 0.5 & 93.3 $\pm$ 1.1 & \textbf{93.7} $\pm$ 0.4 \\
			SUBJ 		&2&  90.3 $\pm$ 4.5 & 91.0 $\pm$ 2.7 & \textbf{93.8} $\pm$ 1.7$^*$ & 83.7 $\pm$ 5.1 & \textbf{94.5} $\pm$ 1.6$^*$ \\
			CB 			&2&  \textbf{88.8} $\pm$ 2.5 & 88.7 $\pm$ 1.9 &  88.0 $\pm$ 3.3 & 83.9 $\pm$ 2.4 & \textbf{89.1} $\pm$ 2.2$^*$ \\
			RTE 		&2&  \textbf{79.9} $\pm$ 1.9 & 79.0 $\pm$ 1.8 &  79.4 $\pm$ 2.1 & 73.8 $\pm$ 3.4 & \textbf{80.6} $\pm$ 1.8$^*$ \\
			AGNews		&4&  88.0 $\pm$ 4.7 & \textbf{89.4} $\pm$ 0.7 & 88.9 $\pm$ 1.3 & 88.0 $\pm$ 0.8 & \textbf{88.8} $\pm$ 1.4 \\
			SST5 		&5&  47.0 $\pm$ 2.6 & \textbf{47.5} $\pm$ 2.3 & 45.0 $\pm$ 2.8 &  \textbf{48.4} $\pm$ 1.0$^*$ & 44.5 $\pm$ 2.4 \\
			TREC 		&6&  87.2 $\pm$ 3.3 & \textbf{90.1} $\pm$ 1.7$^*$ & 88.8 $\pm$ 2.8 & 67.2 $\pm$ 4.8 & \textbf{88.6} $\pm$ 1.7$^*$ \\
			DBPedia 	&14&  88.4 $\pm$ 8.6 & 94.5 $\pm$ 3.0 & \textbf{95.4} $\pm$ 2.6$^*$ & 96.2 $\pm$ 3.0 & \textbf{96.7} $\pm$ 1.4 \\
			NLU Scenario&18&  82.6 $\pm$ 2.0 & \textbf{85.3} $\pm$ 1.5$^*$ & 84.6 $\pm$ 1.7 & 80.2 $\pm$ 2.1 & \textbf{85.8} $\pm$ 1.2$^*$ \\
			TREC Fine	&50&  60.7 $\pm$ 4.8 & \textbf{67.7} $\pm$ 4.3$^*$ & 64.7 $\pm$ 3.7 & 50.1 $\pm$ 4.2 & \textbf{68.6} $\pm$ 4.2$^*$ \\
			NLU Intent 	&68&  68.6 $\pm$ 4.4 & \textbf{74.4} $\pm$ 2.7$^*$ & 60.1 $\pm$ 2.7 &  \textbf{61.6} $\pm$ 3.2 & 61.0 $\pm$ 2.2 \\
			BANKING77	&77&  50.3 $\pm$ 3.1 & \textbf{63.2} $\pm$ 2.5$^*$ & 55.3 $\pm$ 3.5 & 58.1 $\pm$ 2.7 & \textbf{63.7} $\pm$ 3.6$^*$ \\
			CLINIC150	&150&  67.0 $\pm$ 3.6 & \textbf{71.0} $\pm$ 4.2$^*$ & 65.6 $\pm$ 3.0 & 57.2 $\pm$ 2.9 & \textbf{67.3} $\pm$ 2.3$^*$ \\
			\hline
			
		\end{tabular}
	\caption{Comparative Analysis of Classification Accuracy (in \%) for LLAMA-30B Across Various Context Windows (B=3, B=6, B=9). Best scores are highlighted in bold. An asterisk (*) denotes statistical significance, as determined by a t-test with a p-value < 0.05.}
	\label{tab:classification_llama-30b}	
\end{table*}

\begin{table*}[t]
	\centering
	\adjustbox{width=\linewidth}{		
		\begin{tabular}{l|c|ll|ll|ll|ll}  
			\hline					
			\multirow{2}{*}{Dataset}&\multirow{2}{*}{ICL}&\multicolumn{2}{c|}{B=2}&\multicolumn{2}{c|}{B=3} &\multicolumn{2}{c|}{B=4} &\multicolumn{2}{c}{B=6} \\\cline{3-10}
			&& \makecell[c]{PCW}& \makecell[c]{NBCE}& \makecell[c]{PCW}& \makecell[c]{NBCE}& \makecell[c]{PCW}& \makecell[c]{NBCE}& \makecell[c]{PCW}& \makecell[c]{NBCE}\\
			
			\hline

			PIQA &  83.0 $\pm$ 0.6 & \textbf{83.6} $\pm$ 0.6$^*$ & 83.2 $\pm$ 0.6 &  \textbf{83.5} $\pm$ 0.6 & 83.2 $\pm$ 0.7 &  \textbf{83.3} $\pm$ 0.5 & 83.2 $\pm$ 0.6 & 81.9 $\pm$ 1.0 & \textbf{83.2} $\pm$ 0.5$^*$ \\
			OpenBookAQ &  51.0 $\pm$ 1.7 & \textbf{51.1} $\pm$ 1.2$^*$ & 47.0 $\pm$ 1.1 & 50.2 $\pm$ 1.3 & \textbf{50.2} $\pm$ 1.3 & 48.8 $\pm$ 1.1 & \textbf{49.8} $\pm$ 1.0$^*$ & 46.7 $\pm$ 1.3 & \textbf{51.1} $\pm$ 1.0$^*$ \\
			COPA &  79.9 $\pm$ 2.5 & \textbf{81.8} $\pm$ 2.4$^*$ & 79.0 $\pm$ 0.9 &  \textbf{86.0} $\pm$ 1.9$^*$ & 79.8 $\pm$ 2.2 &  \textbf{86.5} $\pm$ 1.5$^*$ & 79.8 $\pm$ 2.1 & 74.9 $\pm$ 3.1 & \textbf{78.4} $\pm$ 1.5$^*$ \\
			HellaSwag &  82.3 $\pm$ 0.7 & \textbf{82.5} $\pm$ 1.0 & 82.5 $\pm$ 0.7 &  \textbf{82.3} $\pm$ 0.7 & 82.2 $\pm$ 0.5 & 82.2 $\pm$ 0.6 & \textbf{82.4} $\pm$ 0.5 & 81.7 $\pm$ 0.8 & \textbf{82.2} $\pm$ 0.5$^*$ \\
			ARCE &  80.3 $\pm$ 0.6 & \textbf{80.5} $\pm$ 0.7$^*$ & 77.4 $\pm$ 0.7 &  \textbf{79.8} $\pm$ 0.5 & 79.7 $\pm$ 0.5 & 78.9 $\pm$ 0.6 & \textbf{79.8} $\pm$ 0.5$^*$ & 76.8 $\pm$ 0.9 & \textbf{80.5} $\pm$ 0.4$^*$ \\
			StoryCloze &  80.5 $\pm$ 0.8 & \textbf{82.1} $\pm$ 0.9$^*$ & 80.1 $\pm$ 0.9 &  \textbf{82.0} $\pm$ 0.6$^*$ & 80.0 $\pm$ 0.9 &  \textbf{81.9} $\pm$ 0.8$^*$ & 80.1 $\pm$ 1.0 &  \textbf{81.2} $\pm$ 0.8$^*$ & 80.1 $\pm$ 0.9 \\
			MMLU &  45.3 $\pm$ 1.8 & \textbf{46.4} $\pm$ 1.9$^*$ & 43.6 $\pm$ 1.3 &  \textbf{45.5} $\pm$ 1.9$^*$ & 44.4 $\pm$ 1.3 &  \textbf{44.7} $\pm$ 2.1 & 44.4 $\pm$ 2.0 & 43.6 $\pm$ 2.8 & \textbf{44.6} $\pm$ 1.4 \\
			
			\hline					
		\end{tabular}
	}			
	\caption{Comparative Results of Task Completion (e.g., Multiple Choices Task)  for LLAMA-13B Using  Various Context Windows. Best scores are highlighted in bold. An asterisk (*) denotes statistical significance, as determined by a t-test with a p-value < 0.05. }
	\label{tab:multi_choice_llama_13b}	
\end{table*}

%
%
%
%
%
%
%
%
%

\begin{table*}[t]
	\centering
	
	\adjustbox{width=\linewidth}{
		\begin{tabular}{l|c|lll|lll}  
			\hline			
			
			\multirow{2}{*}{Dataset}&\multirow{2}{*}{\# Labels}&\multicolumn{3}{c|}{GPT2-Large}&\multicolumn{3}{c}{GPT2-XLarge}  \\\cline{3-8}
			&& \makecell[c]{ICL}& \makecell[c]{PCW}&\makecell[c]{NBCE}& \makecell[c]{ICL}& \makecell[c]{PCW}&\makecell[c]{NBCE}\\
			\hline
			
			SST-2 			&2	&80.2  $\pm$  11.7&84.1  $\pm$  8.2 &\textbf{85.2  $\pm$  6.7} 			        &90.6  $\pm$  3.5  & 92.4 $\pm$  2.5  & \textbf{92.7  $\pm$  2.3*}	 \\			
			CR	 			&2	&81.3  $\pm$  6.3 &81.2  $\pm$  6.4 &\textbf{82.7  $\pm$  6.3} 					&79.2  $\pm$  5.9 &  81.3  $\pm$  4.6&\textbf{82.5  $\pm$  2.9*}	\\
			SUBJ			&2	&65.1  $\pm$  11.9&\textbf{67.0  $\pm$  12.2}&66.1  $\pm$  13.2					&68.8  $\pm$  11.6 & 64.9  $\pm$  7.3 &\textbf{74.5  $\pm$  8.3*}	  \\
			CB 				&2	&43.9  $\pm$  3.7 &43.9  $\pm$  3.2 &\textbf{45.2  $\pm$  3.7} 					&51.9  $\pm$  7.4 &  \textbf{57.2  $\pm$  8.5*}&56.1  $\pm$  7.9	 \\
			RTE				&2	&52.5  $\pm$  2.2 &\textbf{53.5  $\pm$  1.7} &52.9  $\pm$  2.9 					&55.4  $\pm$  2.4  & \textbf{55.6  $\pm$  1.6} &54.9  $\pm$  2.5	\\
			AGNews			&4	&61.7  $\pm$  14.2&70.9  $\pm$  9.4 &\textbf{71.0  $\pm$  8.9 *}			&67.2  $\pm$  13.2 &\textbf{79.6  $\pm$  3.4*} &70.0  $\pm$  9.6	 \\
			SST-5 			&5	&40.8  $\pm$  2.5 &41.5  $\pm$  3.1 &\textbf{41.8  $\pm$  2.4} 				& 38.0  $\pm$  6.1 &  \textbf{41.4  $\pm$  4.3*} &41.1  $\pm$  4.7	  \\
			TREC 			&6	&56.6  $\pm$  7.9 &59.0  $\pm$  4.7 &\textbf{63.1  $\pm$  7.0*}          &47.9 $\pm$  5.1  & 48.7  $\pm$  2.8 &\textbf{51.7  $\pm$  5.0*}	 \\				
			DBPedia			&14	&58.7  $\pm$  20.2&\textbf{78.9  $\pm$  6.6} &71.1  $\pm$  13.7          &77.5 $\pm$  9.8  & 87.0  $\pm$  4.0 &\textbf{87.7  $\pm$  3.8*}	  \\
			
			NLU Scenario	&18	&34.8  $\pm$  7.6 &    28.5  $\pm$  4.3&\textbf{45.7  $\pm$  6.7*}       &45.1  $\pm$  9.3 &    50.0  $\pm$  6.1 & \textbf{51.1  $\pm$  8.1*}  \\ 
			TREC Fine		&50	&36.9  $\pm$  6.3 &    \textbf{37.4  $\pm$  4.8*}&36.9  $\pm$  6.3       &36.4  $\pm$  6.2 &    \textbf{40.1  $\pm$  3.0*} & 40.1  $\pm$  5.1 \\ 
			NLU Intent		&68	&24.5  $\pm$  6.1 &    22.3  $\pm$  5.6&\textbf{27.5  $\pm$  4.6*}       &30.2  $\pm$  5.4 &    33.8  $\pm$  4.6 &\textbf{36.4  $\pm$  4.9*} \\ 
			BANKING77		&77	&28.9  $\pm$  5.1 &    28.0  $\pm$  3.7 &\textbf{36.0  $\pm$  3.2*}      &30.7  $\pm$  4.1 &    33.3  $\pm$  3.5 & \textbf{35.5  $\pm$  2.8*}\\ 
			CLINIC150		&150&43.9  $\pm$  3.2 &    44.1  $\pm$  1.9& \textbf{48.5  $\pm$  2.3*}      &46.6  $\pm$  2.5 &    47.1  $\pm$  2.3 & \textbf{49.9  $\pm$  1.9*} \\
			
			\hline
		\end{tabular}
	}			
	
	\caption{Comparative analysis of classification results in terms of accuracy (in \%) for both the GPT2-Large and GPT2-XLarge models using a context window of B = 3. Notably, a single window comprises a set of examples with a total number of tokens equal to the maximum capacity of conventional in-context learning (e.g., 1024 tokens in GPT-2). The best-performing scores for each model and dataset are highlighted in bold, while '*' indicates statistical significance, determined by a t-test with a p-value < 0.05.}
	\label{tab:classification_gpt}	
\end{table*}	

\begin{table*}
	\centering
	
	\adjustbox{width=\linewidth}{
		\begin{tabular}{l|c|lll|lll|lll}  
			\hline			
			
			\multirow{2}{*}{Dataset}&\multirow{2}{*}{\# Labels}&\multicolumn{3}{c|}{OPT-1.3B}&\multicolumn{3}{c}{OPT-6.7B}&\multicolumn{3}{c}{OPT-13B} \\\cline{3-11}
			&&   \makecell[c]{ICL}& \makecell[c]{PCW}&\makecell[c]{NBCE}& \makecell[c]{ICL}& \makecell[c]{PCW}&\makecell[c]{NBCE}& \makecell[c]{ICL}& \makecell[c]{PCW}&\makecell[c]{NBCE}\\
			
			\hline
			
			SST-2 			&2	& 85.0  $\pm$  8.5 			&  81.7  $\pm$  10.6&\textbf{86.0  $\pm$  7.2}  		& 93.8  $\pm$  2.6 &  93.7  $\pm$  3.3&\textbf{95.8  $\pm$  1.7*} 	&93.1  $\pm$  4.4 & 93.8  $\pm$  3.1&\textbf{94.9  $\pm$  2.3}\\			
			CR	 	        &2  & 89.1  $\pm$  2.4 			&  88.8  $\pm$  2.3 &\textbf{89.7  $\pm$  1.7}    		&90.3  $\pm$  2.5 &90.7  $\pm$  2.4 &\textbf{91.7  $\pm$  1.5*} 	&92.7  $\pm$  1.5 & 92.3  $\pm$  2.5&\textbf{93.1  $\pm$  1.4}\\
			SUBJ			&2	& \textbf{78.8  $\pm$  9.0*}&  68.3  $\pm$  7.5 &69.0  $\pm$  7.9   				&\textbf{72.3  $\pm$  10.6*}&70.9  $\pm$  13.9&64.0  $\pm$  10.7   &86.4  $\pm$  9.2 & 88.0  $\pm$  8.3&\textbf{90.1  $\pm$  5.9}\\
			CB 				&2	& \textbf{53.0  $\pm$  6.0} &  50.5  $\pm$ 3.3 	&50.8  $\pm$  3.3   				&52.4  $\pm$  10.1&\textbf{59.9  $\pm$  12.1}&59.3  $\pm$  10.8 		&50.5  $\pm$  8.5 & 49.3  $\pm$  5.8&\textbf{62.5  $\pm$  10.2}\\
			RTE				&2	&  51.1  $\pm$  3.7 		&  51.8  $\pm$  3.8 &\textbf{52.7  $\pm$  3.2} 			&56.1  $\pm$  2.2 &56.2  $\pm$  1.6 &\textbf{56.8  $\pm$  2.0} 	&53.0  $\pm$  6.0 & 56.3  $\pm$  4.9&\textbf{56.8  $\pm$  6.2}\\
			AGNews			&4	&  61.3  $\pm$  10.3		& \textbf{67.4  $\pm$  6.7*} &59.6  $\pm$  7.2      	&74.8  $\pm$  6.7 &\textbf{76.7  $\pm$  4.8*} &72.7  $\pm$  5.7 	&78.6  $\pm$  5.6 &\textbf{82.4  $\pm$  2.3}&78.8  $\pm$  3.9\\
			SST-5 			&5	& 44.0  $\pm$  3.9 			&  42.7  $\pm$  4.6 &\textbf{44.8  $\pm$  2.8}     		&42.7  $\pm$  5.1 &  \textbf{45.2  $\pm$  4.2} &42.5  $\pm$  4.6  	&45.6  $\pm$  3.4 & \textbf{45.7  $\pm$  2.6}&42.9  $\pm$  4.2\\
			TREC 			&6	&\textbf{59.4  $\pm$  6.3* }&  55.0  $\pm$ 4.3 	&56.8  $\pm$  4.7    				&70.3  $\pm$  3.3 &\textbf{73.1  $\pm$  2.2*} &71.8  $\pm$  3.5    	&56.7  $\pm$  7.2 &\textbf{62.4  $\pm$  6.2} &57.1  $\pm$  6.8\\				
			DBPedia			&14	&  86.3  $\pm$  3.8 		&  87.7  $\pm$  2.1 &\textbf{87.9  $\pm$  2.2}      	&89.8  $\pm$ 3.5  &\textbf{94.3  $\pm$  2.0*} &93.5  $\pm$  2.6  	&87.3  $\pm$  4.0 & \textbf{94.1  $\pm$  2.1}&94.0  $\pm$  2.2\\
			
			NLU Scenario	&18	&67.8  $\pm$  4.0 			& 69.9  $\pm$  3.5	&  \textbf{70.2  $\pm$  4.0}  		&74.9  $\pm$  3.0 &\textbf{79.0  $\pm$  2.0}&77.9  $\pm$  3.0&78.5  $\pm$  3.2 & {81.8  $\pm$  2.0}     &\textbf{83.7  $\pm$  1.8}   \\ 
			TREC Fine		&50	&39.7  $\pm$  4.5 			& 38.8  $\pm$  4.7	& \textbf{41.5  $\pm$  6.0}   		&45.7  $\pm$  6.7 & 49.6  $\pm$  6.6 &\textbf{50.1  $\pm$  6.7}&49.7  $\pm$  6.0 &\textbf{55.5  $\pm$  6.6}           &  51.7  $\pm$  6.6  \\ 
			NLU Intent		&68	&45.3  $\pm$  4.9 			& 50.0  $\pm$  4.2	&  \textbf{50.9  $\pm$  4.0}   		&55.8  $\pm$  3.9 & 62.5  $\pm$  3.1&\textbf{63.3  $\pm$  3.1}&61.5  $\pm$  2.8 &\textbf{71.8  $\pm$  2.5}           & \textbf{71.8    $\pm$  2.7}    \\ 
			BANKING77		&77	&25.9  $\pm$  4.9 			& 24.8  $\pm$  4.0	&  \textbf{28.8  $\pm$  4.5}   		&43.6  $\pm$  3.1&51.9  $\pm$  2.8&\textbf{53.7  $\pm$  3.3} &43.3  $\pm$  3.4 & 53.0  $\pm$  3.8          & \textbf{56.0    $\pm$  3.4 }    \\ 
			CLINIC150		&150&50.8  $\pm$  3.0 			& 52.4  $\pm$  2.3	& \textbf{57.7  $\pm$  2.0} 		&60.4  $\pm$  2.4 & 63.0  $\pm$  1.9 &\textbf{65.5  $\pm$  1.9} &59.7  $\pm$  2.3 & 65.1  $\pm$  2.7       & \textbf{66.1  $\pm$  2.1} \\
			
			\hline			
		\end{tabular}
	}			
	
	\caption{Comparative analysis of classification results measured by accuracy (in \%) for OPT models with B = 3.  The best scores are highlighted in bold, while '*' indicates p-value < 0.05.}
	\label{tab:classification_opt}	
\end{table*}

\begin{table*}[t]
	\centering
	
	\adjustbox{width=\linewidth}{
		\begin{tabular}{l|c|c|ll|ll|ll}  
			\hline					
			\multirow{2}{*}{Dataset}&\multirow{2}{*}{\# Labels}&\multirow{2}{*}{ICL}&\multicolumn{2}{c|}{B=3}&\multicolumn{2}{c|}{B=4} &\multicolumn{2}{c}{B=5} \\\cline{4-9}
			&&& \makecell[c]{PCW}& \makecell[c]{NBCE}& \makecell[c]{PCW}& \makecell[c]{NBCE}& \makecell[c]{PCW}& \makecell[c]{NBCE}\\
			
			\hline			
			SST-2 			&2	& 93.8  $\pm$  2.6 &  93.7  $\pm$  3.3&\textbf{95.8  $\pm$  1.7*} 	 &93.9  $\pm$  2.7	& \textbf{ 96.1  $\pm$  0.9* }    &92.3  $\pm$  4.2  &  \textbf{96.3  $\pm$  0.9*}  \\			 
			CR	 			&2	&90.3  $\pm$  2.5 &90.7  $\pm$  2.4 &\textbf{91.7  $\pm$  1.5*} 	&90.8  $\pm$  2.3	&  \textbf{91.9  $\pm$  1.6* }  	&90.0  $\pm$  2.7  &  \textbf{91.5  $\pm$  1.4*}  \\
			SUBJ			&2	&\textbf{72.3  $\pm$  10.6*}&70.9  $\pm$  13.9&64.0  $\pm$  10.7    &\textbf{66.6  $\pm$  13.2	}&  65.7  $\pm$  9.7&67.3  $\pm$  14.2 &  \textbf{68.4  $\pm$  9.8}  \\
			CB 				&2	&52.4  $\pm$  10.1&\textbf{59.9  $\pm$  12.1}&59.3  $\pm$  10.8 		 &55.6  $\pm$  10.4	& \textbf{59.8  $\pm$  12.0}     	&\textbf{60.7  $\pm$  8.7}  &  56.1  $\pm$  9.9  \\
			RTE				&2	&56.1  $\pm$  2.2 &56.2  $\pm$  1.6 &\textbf{56.8  $\pm$  2.0} 	 &55.7  $\pm$  1.6	&  \textbf{56.6  $\pm$  2.0 }	        &55.0  $\pm$  1.4  &  \textbf{56.9  $\pm$  1.9*}  \\
			AGNews			&4	&74.8  $\pm$  6.7 &\textbf{76.7  $\pm$  4.8*} &72.7  $\pm$  5.7 	 &75.7  $\pm$  5.3	& \textbf{73.0  $\pm$  5.6}       	&\textbf{77.7  $\pm$  3.9}  &  77.1  $\pm$  5.1  \\
			SST-5 			&5	&42.7  $\pm$  5.1 &  \textbf{45.2  $\pm$  4.2} &42.5  $\pm$  4.6  	 &\textbf{44.3  $\pm$  4.5*}	&  41.3  $\pm$  3.5 	        &\textbf{46.3  $\pm$  3.6*}  &  42.8  $\pm$  3.4  \\
			TREC 			&6	&70.3  $\pm$  3.3 &\textbf{73.1  $\pm$  2.2*} &71.8  $\pm$  3.5    	  & \textbf{72.1  $\pm$  2.9}	&  72.0  $\pm$  3.4 	       &\textbf{73.6  $\pm$  2.7}  &  72.9  $\pm$  2.9  \\				
			DBPedia			&14&89.8  $\pm$ 3.5  &\textbf{94.3  $\pm$  2.0*} &93.5  $\pm$  2.6  	& \textbf{94.4  $\pm$  2.1} &  93.4  $\pm$  2.3 	        &\textbf{94.7  $\pm$  1.5*}  &  93.7  $\pm$  2.0  \\
			
			NLU Scenario	&18&74.9  $\pm$  3.0 &\textbf{79.0  $\pm$  2.0}&77.9  $\pm$  3.0	& \textbf{76.8  $\pm$  4.3*}& \textbf{76.8  $\pm$  3.1*} &77.7  $\pm$  3.8  &  \textbf{79.3  $\pm$  2.1*} \\ 
			TREC Fine		&50&45.7  $\pm$  6.7 & 49.6  $\pm$  6.6 &\textbf{50.1  $\pm$  6.7}	&48.2  $\pm$  6.7&  \textbf{49.4  $\pm$  6.9} &\textbf{51.5  $\pm$  6.9}  &  {50.7  $\pm$  5.2 }\\ 
			NLU Intent		&68&55.8  $\pm$  3.9 & 62.5  $\pm$  3.1&\textbf{63.3  $\pm$  3.1}	&61.8  $\pm$  3.6&  \textbf{62.4  $\pm$  3.9} &61.1  $\pm$  3.7  & \textbf{66.4  $\pm$  2.3* }\\ 
			BANKING77		&77	&43.6  $\pm$  3.1&51.9  $\pm$  2.8&\textbf{53.7  $\pm$  3.3} 	&51.5  $\pm$  3.2&  53.8  $\pm$  3.2 &52.2  $\pm$  2.0  &  \textbf{56.4  $\pm$  2.6} \\ 
			CLINIC150		&150&60.4  $\pm$  2.4 & 63.0  $\pm$  1.9 &\textbf{65.5  $\pm$  1.9} &62.7  $\pm$  2.2&  \textbf{65.5  $\pm$  2.5*} &61.9  $\pm$  1.8  & \textbf{67.1  $\pm$  2.2*}\\
			
			%

			\hline
		\end{tabular}
	}			
	
	\caption{The comparative results of context extension, measured by accuracy (in \%), for OPT-6.7B models with windows (B = 4 and B = 5). }
	\label{tab:opt_6.7b}	
\end{table*}

\begin{table*}[t]
	\centering
	
	\adjustbox{width=\linewidth}{
		\begin{tabular}{l|c|ll|ll|ll|ll}  
			\hline								
			\multirow{3}{*}{Dataset}&\multirow{3}{*}{\# Labels}&\multicolumn{4}{c|}{GPT2-Large}&\multicolumn{4}{c}{GPT2-XLarge}\\\cline{3-10}		
			&&\multicolumn{2}{c|}{B = 4}&\multicolumn{2}{c|}{B = 5}&\multicolumn{2}{c|}{B = 4}&\multicolumn{2}{c}{B = 5}\\\cline{3-10}					
			&& \makecell[c]{PCW}&\makecell[c]{NBCE} & \makecell[c]{PCW}&\makecell[c]{NBCE} & \makecell[c]{PCW}&\makecell[c]{NBCE} & \makecell[c]{PCW}&\makecell[c]{NBCE} \\

			\hline			
			SST-2 			&2	&83.3  $\pm$  7.8  	&  \textbf{83.9  $\pm$  7.9 }&\textbf{85.0  $\pm$  6.9}  &  83.7  $\pm$  8.6 		&91.3  $\pm$  2.9&  \textbf{92.6  $\pm$  2.6} &91.4  $\pm$  3.1  &  \textbf{92.4  $\pm$  2.4}  \\			
			CR	 			&2	&82.1  $\pm$  5.9  	&  \textbf{84.1  $\pm$  5.7} &81.7  $\pm$  4.7  &  \textbf{82.4  $\pm$  5.1}		&82.1  $\pm$  2.9&  \textbf{82.7  $\pm$  3.0} &\textbf{82.0  $\pm$  2.4}  &  81.7  $\pm$  2.5  \\
			SUBJ			&2	&\textbf{68.1  $\pm$  11.9}  &  63.1  $\pm$  10.5&66.5  $\pm$  10.3  &  \textbf{68.9  $\pm$  10.5}   	&63.9  $\pm$  6.0&  \textbf{76.2  $\pm$  6.7} &59.3  $\pm$  5.2  & \textbf{79.3  $\pm$  5.5*}  \\
			CB 				&2	&44.0  $\pm$  3.4  	&  \textbf{44.7  $\pm$  4.3} &42.8  $\pm$  2.0  &  \textbf{43.8  $\pm$  2.8}  	&\textbf{53.9  $\pm$  6.2}&  53.8  $\pm$  9.1 &51.1  $\pm$  4.4  &  \textbf{56.7  $\pm$  7.7*}  \\
			RTE				&2	&\textbf{53.5  $\pm$  1.5*}  &  52.1  $\pm$  3.0 &54.0  $\pm$  1.2  &  53.7  $\pm$  2.2  	&\textbf{55.3  $\pm$  1.1}&  54.7  $\pm$  3.0 &54.9  $\pm$  1.7  &  \textbf{55.7  $\pm$  1.7}  \\
			AGNews			&4	&\textbf{69.2  $\pm$  9.6}  	&  68.1  $\pm$  12.5&67.9  $\pm$  8.1  &  \textbf{70.7  $\pm$  8.4}  	&\textbf{80.5  $\pm$  3.3*}&  72.5  $\pm$  8.8 &\textbf{80.0  $\pm$  2.5*}  &  73.0  $\pm$  6.7  \\
			SST-5 			&5	&40.1  $\pm$  4.0  	&  \textbf{42.4  $\pm$  1.7*} &40.4  $\pm$  3.9  & \textbf{42.6  $\pm$  1.6}  	&\textbf{41.5  $\pm$  4.2*}&  38.5  $\pm$  5.7 &39.2  $\pm$  4.4  &  \textbf{41.7  $\pm$  5.8*} \\
			TREC 			&6	&57.4  $\pm$  4.1	&  \textbf{64.8  $\pm$  4.0*} &55.3  $\pm$  4.0  &  \textbf{64.6  $\pm$  4.8*}   	&48.9  $\pm$  3.4&  \textbf{51.6  $\pm$  3.7} &48.1  $\pm$  2.2  &  \textbf{53.0  $\pm$  2.7*}  \\				
			DBPedia			&14	&\textbf{80.7  $\pm$  5.0*}  &  74.8  $\pm$  12.1&\textbf{79.3  $\pm$  4.4}  &  76.5  $\pm$  8.4  	&\textbf{88.5  $\pm$  3.3}&  87.5  $\pm$  4.7 &\textbf{89.8  $\pm$  3.2}  &  89.1  $\pm$  3.6  \\
			
			NLU Scenario	&18	&27.8  $\pm$  3.6 	&  \textbf{46.6  $\pm$  7.4} &27.5  $\pm$  3.3  &  \textbf{44.4  $\pm$  6.5}  	&49.7  $\pm$  5.7&  \textbf{51.7  $\pm$  7.6} &48.7  $\pm$  6.0  &  \textbf{52.8  $\pm$  5.5*}  \\ 
			TREC Fine		&50	&32.4  $\pm$  5.1 	&  \textbf{37.4  $\pm$  4.8*} &31.2  $\pm$  4.1  & \textbf{ 39.9  $\pm$  3.6*}  	&38.6  $\pm$  3.1&  \textbf{39.8  $\pm$  6.1} &37.2  $\pm$  2.3  &  \textbf{41.6  $\pm$  3.8*}  \\ 
			NLU Intent		&68	&24.3  $\pm$  4.7	&  \textbf{26.0  $\pm$  5.6} &20.3  $\pm$  5.4  &  \textbf{27.3  $\pm$  4.4}  	&34.8  $\pm$  5.1 &  \textbf{35.9  $\pm$  5.2} &  37.1  $\pm$  5.1 &  \textbf{38.6  $\pm$  3.3*}  \\ 
			BANKING77		&77	&26.6  $\pm$  3.2	&  \textbf{35.2  $\pm$  3.8*} &25.5  $\pm$  3.2  &  \textbf{36.0  $\pm$  3.8*}  	&31.0  $\pm$  3.5&  \textbf{35.4  $\pm$  3.2*}&29.6  $\pm$  2.8  &  \textbf{37.7  $\pm$  2.6*}  \\ 
			CLINIC150		&150&43.2  $\pm$  1.8	&  \textbf{48.1  $\pm$  1.9*} &41.6  $\pm$  2.2  & \textbf{49.4  $\pm$  2.0*} 	&45.9  $\pm$  2.9&  \textbf{49.3  $\pm$  2.3*}&43.0  $\pm$  2.4  &  \textbf{50.3  $\pm$  2.5*} \\
			\hline
		\end{tabular}
	}		
	
	\caption{The comparative results of classification tasks, quantified in terms of accuracy (in \%), for both GPT2-Large and GPT2-XLarge models using different context windows (B = 4 and B = 5). The best scores for each model and dataset are highlighted in bold, while an asterisk (*) denotes statistical significance (as determined by a t-test with a p-value < 0.05).
	}
	\label{tab:gpt_more_b}	
\end{table*}

\begin{table*}[t]
	\centering
	
	\adjustbox{width=\linewidth}{
		\begin{tabular}{l|c|ll|ll|ll|ll}  
			\hline						
			
			\multirow{3}{*}{Dataset}&\multirow{3}{*}{\# Labels}&\multicolumn{4}{c|}{OPT-1.3B}&\multicolumn{4}{c}{OPT-6.7B}\\\cline{3-10}		
			&&\multicolumn{2}{c|}{B = 4}&\multicolumn{2}{c|}{B = 5}&\multicolumn{2}{c|}{B = 4}&\multicolumn{2}{c}{B = 5}\\\cline{3-10}					
			&& \makecell[c]{PCW}&\makecell[c]{NBCE}& \makecell[c]{PCW}&\makecell[c]{NBCE}& \makecell[c]{PCW}&\makecell[c]{NBCE}& \makecell[c]{PCW}&\makecell[c]{NBCE}\\
			\hline			
			SST-2 			&2	&81.1  $\pm$  7.7			&  \textbf{88.1  $\pm$  5.7*} 	& 79.9  $\pm$  9.8      	&  \textbf{88.8  $\pm$  5.2*} 	&93.9  $\pm$  2.7	& \textbf{ 96.1  $\pm$  0.9* }    &92.3  $\pm$  4.2  &  \textbf{96.3  $\pm$  0.9*}  \\			
			CR	 			&2	&88.5  $\pm$  3.3			&  \textbf{88.8  $\pm$  1.6} 	& 85.6  $\pm$  3.6      	&  \textbf{89.1  $\pm$  1.5*} 	&90.8  $\pm$  2.3	&  \textbf{91.9  $\pm$  1.6* }  	&90.0  $\pm$  2.7  &  \textbf{91.5  $\pm$  1.4*}  \\
			SUBJ			&2	&68.5  $\pm$  6.6			& \textbf{70.5  $\pm$  7.4 }	& 65.2  $\pm$  8.3       	&  \textbf{70.9  $\pm$  6.3*} 	&\textbf{66.6  $\pm$  13.2	}&  65.7  $\pm$  9.7&67.3  $\pm$  14.2 &  \textbf{68.4  $\pm$  9.8}  \\
			CB 				&2	&\textbf{51.6  $\pm$  5.2}	&  51.5  $\pm$  4.3 			& 49.1  $\pm$  1.0     		&  \textbf{51.6  $\pm$  3.6*} 	&55.6  $\pm$  10.4	& \textbf{59.8  $\pm$  12.0}     	&\textbf{60.7  $\pm$  8.7}  &  56.1  $\pm$  9.9  \\
			RTE				&2	&50.6  $\pm$  3.1			&  \textbf{51.4  $\pm$  2.9 }	& 50.9  $\pm$  2.1      	&  \textbf{51.3  $\pm$  2.5} 	&55.7  $\pm$  1.6	&  \textbf{56.6  $\pm$  2.0 }	        &55.0  $\pm$  1.4  &  \textbf{56.9  $\pm$  1.9*}  \\
			AGNews			&4	&\textbf{65.1  $\pm$  5.9*}	&  60.3  $\pm$  9.0 			&\textbf{ 69.4  $\pm$  5.0*}&  {62.9  $\pm$  6.7} 	&75.7  $\pm$  5.3	& \textbf{73.0  $\pm$  5.6}       	&\textbf{77.7  $\pm$  3.9}  &  77.1  $\pm$  5.1  \\
			SST-5 			&5	&42.4  $\pm$  4.0			& \textbf{44.8  $\pm$  2.2*} 	& 41.6  $\pm$  4.3          &\textbf{45.1  $\pm$  2.0*}   	&\textbf{44.3  $\pm$  4.5*}	&  41.3  $\pm$  3.5 	        &\textbf{46.3  $\pm$  3.6*}  &  42.8  $\pm$  3.4  \\
			TREC 			&6	&55.2  $\pm$  3.2			&  \textbf{55.7  $\pm$  4.3} 	& 52.5  $\pm$  2.8          &  \textbf{57.1  $\pm$  3.9*} 	&\textbf{72.1  $\pm$  2.9}	&  72.0  $\pm$  3.4 	        &\textbf{73.6  $\pm$  2.7}  &  72.9  $\pm$  2.9  \\				
			DBPedia			&14	&\textbf{88.1  $\pm$  2.6}	&  87.5  $\pm$  2.6 			& 87.0  $\pm$  3.1          &  \textbf{87.9  $\pm$  2.6} 	&\textbf{94.4  $\pm$  2.1}	&  93.4  $\pm$  2.3 	        &\textbf{94.7  $\pm$  1.5*}  &  93.7  $\pm$  2.0  \\
			
			NLU Scenario	&18	&\textbf{69.9  $\pm$  2.6}	&  69.3  $\pm$  4.3 			&67.7  $\pm$  4.0           &  \textbf{72.8  $\pm$  3.8*} 	&\textbf{76.8  $\pm$  4.3*}& \textbf{76.8  $\pm$  3.1*} &77.7  $\pm$  3.8  &  \textbf{79.3  $\pm$  2.1*} \\ 
			TREC Fine		&50	&40.5  $\pm$  5.8			&\textbf{43.1  $\pm$  6.4} 		&35.3  $\pm$  3.5           &  \textbf{42.0  $\pm$  4.7*} 	&48.2  $\pm$  6.7&  \textbf{49.4  $\pm$  6.9} &\textbf{51.5  $\pm$  6.9}  &  {50.7  $\pm$  5.2 }\\ 
			NLU Intent		&68	&48.8  $\pm$  4.2			& \textbf{51.0  $\pm$  4.7} 	&45.4  $\pm$  3.2           &  \textbf{54.5  $\pm$  3.3*} 	&61.8  $\pm$  3.6&  \textbf{62.4  $\pm$  3.9} &61.1  $\pm$  3.7  & \textbf{66.4  $\pm$  2.3* }\\ 
			BANKING77		&77	&26.0  $\pm$  3.5			&\textbf{30.1  $\pm$  3.5*} 	&28.9  $\pm$  3.1           &  \textbf{32.5  $\pm$  3.5*} 	&51.5  $\pm$  3.2&  53.8  $\pm$  3.2 &52.2  $\pm$  2.0  &  \textbf{56.4  $\pm$  2.6} \\ 
			CLINIC150		&150&52.6  $\pm$  2.0			&  \textbf{57.2  $\pm$  2.5* }	&49.3  $\pm$  2.5           &  \textbf{58.4  $\pm$  2.0*}	&62.7  $\pm$  2.2&  \textbf{65.5  $\pm$  2.5*} &61.9  $\pm$  1.8  & \textbf{67.1  $\pm$  2.2*}\\
			
			%

			\hline
		\end{tabular}
	}			
	
	\caption{The comparative results of context extension, measured by accuracy (in \%), for OPT models with windows (B = 4 and B = 5). }
	\label{tab:opt_more_b}	
\end{table*}

\subsection{Prompt Format} \label{sec:prompt}

\begin{table*}[ht]
	\centering

	\adjustbox{width=\linewidth}{		
		\begin{tabular}{l|cc|l|p{6cm}}
			\hline
			\textbf{Dataset} &\multicolumn{2}{c|}{\makecell{Number of shots per window B}}  & \multirow{2}{*}{Prompt Example} & \multirow{2}{*}{Labels} \\
			&\textbf{$k_{max}$ GPT2} & \textbf{$k_{max}$ LLAMA}&&\\
			\hline

			SST-2 & 27 & 48 & \makecell[l]{Sentence: \{Sentence\}\\Label: {Label}} & [negative, positive] \\
			\hline
			
			CR	&21&39&\makecell[l]{Review:\{Sentence\}\\ Sentiment:\{Label\}}&[negative, positive]\\
			\hline 		
			SUBJ	&18&32&\makecell[l]{Input:\{Sentence\}\\ Type:\{Label\}}&[objective, subjective]\\
			\hline 			
			CB 		&5&10&\makecell[l]{Premise:\{Sentence\}\\Hypothesis:\{\ hypothesis\}\\ Prediction:\{Label\}}&[true, false, neither]\\
			\hline 		
			RTE	&5&10&\makecell[l]{Premise:\{Sentence\}\\Hypothesis:\{\ hypothesis\}\\ Prediction:\{Label\}}&[True, False]\\
			\hline 	
			AGNews	&11&20&\makecell[l]{Input:\{Sentence\}\\ Type:\{Label\}}&[world, sports, business, technology]\\
			\hline 		
			SST-5&20&36&\makecell[l]{Review:\{Sentence\}\\ Sentiment:{{{Sentiment}}}}&[terrible, bad, okay, good, great]\\
			\hline 	
			TREC 	&38&69&\makecell[l]{Question:\{Sentence\}\\ Type:\{Label\}}&[abbreviation, entity, description, human, location, numeric]\\
			\hline 							
			DBPedia		&7&14&\makecell[l]{Input:\{Sentence\}\\ Type:\{Label\}}&[company, school, artist, athlete, politics, transportation, building, nature, village, animal, plant,
			album, film, book]\\
			\hline 	
			NLU Scenario&43&80&\makecell[l]{Utterance:\{Sentence\}\\ Scenario:\{Label\}}&[lists, weather, general, cooking, email, alarm,
			datetime, calendar, social, transport, iot, recommendation, takeaway, play, music, qa, news, audio]\\
			\hline 	
			TREC Fine	&37&65&\makecell[l]{Question:\{Sentence\}\\ Type:\{Label\}}&[abbreviation abbreviation, abbreviation expansion, entity animal, entity body, entity color, entity creation, entity currency, entity disease, entity
			event, entity food...\\
			\hline 	
			NLU Intent	&43&80&\makecell[l]{Utterance:\{Sentence\}\\ Intent:\{Label\}}&[alarm query, alarm remove, alarm set, audio
			volume down, audio volume mute, audio volume
			other, audio volume up, calendar query, calendar
			remove, calendar set...\\
			\hline 	
			BANKING77&27&51&\makecell[l]{Query:{\{Sentence\}}\\ Intent:\{Label\}}&[activate my card, age limit, apple pay or google
			pay, atm support, automatic top up, balance not
			updated after bank transfer, balance not updated
			after cheque or cash deposit...\\
			\hline 	
			CLINIC150&39&72&\makecell[l]{Sentence:\{Sentence\}\\ Intent:\{Label\}}&[restaurant reviews, nutrition info, account
			blocked, oil change how, time, weather, redeem
			rewards, interest rate, gas type...\\
			\hline
		\end{tabular}
	}
	
	\caption{Classification datasets with used prompts and $k_{max}$ for GPT2 and LLaMA. Note that OPT shares the same length of LLAMA (i.e., 2048)}
	\label{tab:format}
\end{table*}

\end{document}